\PassOptionsToPackage{numbers}{natbib}
\PassOptionsToPackage{sort}{natbib}

\documentclass{article}

\usepackage{geometry}
\usepackage[utf8]{inputenc} 
\usepackage[T1]{fontenc}   
\usepackage{fullpage}
\usepackage[numbers]{natbib}

\date{}

\usepackage[pdfencoding=auto, psdextra]{hyperref}
\usepackage{url}
\usepackage{microtype}
\usepackage{graphicx}
\usepackage{subcaption}
\usepackage{booktabs} % for professional tables

\usepackage{ graphicx }
\usepackage{ subcaption}
\usepackage{ mathtools }
\usepackage{ multirow }
\usepackage{ tablefootnote }
\usepackage{ xcolor } 
\usepackage{ nicefrac }
\usepackage{ amsfonts }
\usepackage{ url }
\usepackage{ caption }
\usepackage{ wrapfig }
\usepackage[outline]{ contour }% http://ctan.org/pkg/contour
\usepackage{ makecell }
\usepackage{ xcolor }
\usepackage[T1]{ fontenc }
\usepackage{ amsmath }
\usepackage{ amssymb }
\usepackage{ mathtools }
\usepackage{ amsthm }
\usepackage{ amssymb }
\usepackage{ bbm}
\usepackage{ wasysym }
\usepackage{ esvect }
\usepackage{ listings }
\usepackage[frozencache,cachedir=minted]{ minted }
\usepackage{ multirow }
\usepackage{ multicol }
\usepackage{ algorithm }
\usepackage{ hyperref }
\usepackage{ other_commands }
\usepackage{ math_commands }

\definecolor{grey}{rgb}{0.5,0.5,0.5}

\newcommand{\x}{\rvx}
\newcommand{\y}{\ry}
\newcommand{\w}{\vw}
\newcommand{\pw}{p_\vw}
\newcommand{\hD}{\hat{\calD}}
\newcommand{\xone}{\ervx_1}
\newcommand{\xtwo}{\rvx_2}
\newcommand{\wone}{\evw_1}
\newcommand{\wtwo}{\vw_2}
\newcommand{\wtx}{\innerprod{\vw}{\rvx}}
\newcommand{\distr}{\gD}
\newcommand{\empdistr}{{\hat \gD }}
\newcommand{\tSigma}{\tilde{\Sigma}}

\newcommand{\sigmaMintSigma}{\sigma_{\mathrm{min}}(\tSigma)}
\newcommand{\aw}{a_\vw}

\newcommand{\ltwonorm}[1]{\norm{#1}_2}

\newcommand{\erfc}[1]{\mathrm{erfc}\paren{#1}}
\newcommand{\erfcinv}[1]{\mathrm{erfc}^{-1}\paren{#1}}

\newcommand{\sigmaw}{\sigma_\vw}
\newcommand{\wzero}{{\vw^{(0)}}}
\newcommand{\witer}[1]{{\vw^{(#1)}}}
\newcommand{\woneiter}[1]{{\evw^{(#1)}_1}}
\newcommand{\wtildeoneiter}[1]{{\tilde{\evw}^{(#1)}_1}}
\newcommand{\wtwoiter}[1]{{\vw^{(#1)}_2}}
\newcommand{\wtildetwoiter}[1]{{\tilde{\vw}^{(#1)}_2}}
\newcommand{\marginlosswxy}{l_\gamma(\vw; (\rvx, \ry))}
\newcommand{\errD}{\E_\calD \brck{ \I( \ry \cdot \innerprod{\vw}{\rvx} < 0)}}
\newcommand{\errDhat}{\E_{\hat \calD} \brck{ \I( \ry \cdot \innerprod{\vw}{\rvx} < 0)}}
\newcommand{\marginerrDhat}{\E_{\hat \calD} \brck{  \I (\ry \cdot \innerprod{\vw}{\rvx} <\gamma)}}
\newcommand{\opnorm}[1]{\|#1\|_{\mathrm{op}}}
\newcommand{\pardev}[2]{\frac{\partial #1}{\partial #2}}

\renewcommand{\leq}{\leqslant}
\renewcommand{\geq}{\geqslant}

\renewcommand*{\eqref}[1]{Equation~\ref{#1}} 

\title{Adversarial Unlearning: \\ Reducing Confidence Along Adversarial Directions}

\author{%
Amrith Setlur$^{1,}$\thanks{Correspondence can be sent to \href{mailto:asetlur@cs.cmu.edu}{asetlur@cs.cmu.edu}.} \qquad Benjamin Eysenbach$^{1}$ \qquad Virginia Smith$^{1}$ \qquad Sergey Levine$^{2}$ \\
    \small $^1$ Carnegie Mellon University \quad $^2$ UC Berkeley \\
}

\begin{document}

\maketitle

\begin{abstract}
Supervised learning methods trained with maximum likelihood objectives often overfit on training data. Most regularizers that prevent overfitting look to increase confidence on additional examples (e.g., data augmentation, adversarial training), or reduce it on training data (e.g., label smoothing). In this work we propose a complementary regularization strategy that reduces confidence on self-generated examples. The method, which we call RCAD (Reducing Confidence along Adversarial Directions), aims to reduce confidence on out-of-distribution examples lying along directions adversarially chosen to increase training loss. In contrast to adversarial training, RCAD does not try to robustify the model to output the original label, but rather regularizes it to have reduced confidence on points generated using much larger perturbations than in conventional adversarial training. RCAD can be easily integrated into training pipelines with a few lines of code. Despite its simplicity, we find on many classification benchmarks that RCAD can be added to existing techniques (e.g., label smoothing, MixUp training) to increase test accuracy by 1--3$\%$ in absolute value, with more significant gains in the low data regime. We also provide a theoretical analysis that helps to explain these benefits in simplified settings, showing that RCAD can provably help the model unlearn spurious features in the training data.  
\end{abstract}

\section{Introduction}
\label{sec:introduction}

Supervised learning techniques typically consider training models to make accurate predictions on fresh test examples drawn from the same distribution as training data.
Unfortunately, it is well known that maximizing the likelihood of the training data alone may result in overfitting. Prior work broadly considers two approaches to combat this issue. Some methods train  on additional examples, \eg generated via augmentations~\citep{shorten2019survey,devries2017improved,yun2019cutmix} or adversarial updates~\citep{goodfellow2014explaining,madry2017towards,carlini2019evaluating}. 
Others modify the objective by using alternative losses and/or regularization terms (\eg label smoothing \citep{szegedy2016rethinking,muller2019does}, MixUp~\citep{zhang2018mixup}, robust objectives~\citep{wang2013robust,hertz2018introduction}).
In effect, these prior approaches either make the model's predictions more certain on new training examples or make the distribution over potential models less certain.

Existing regularization methods can be seen as providing a certain inductive bias for the model, e.g., the model's weights should be small (\ie weight decay), the model's predictions should vary linearly between training examples (\ie MixUp). In this paper we identify a different inductive bias: the model's predictions should be less confident on out-of-distribution inputs that look nothing like the training examples. We turn this form of inductive bias into a simple regularizer, whose benefits are complementary to existing regularization strategies.
To instantiate such a method, we must be able to sample out-of-distribution examples. For this, we propose a simple approach: generating adversarial examples~\citep{moosavi2016deepfool,goodfellow2014explaining} using very large step sizes (orders-of-magnitude larger than traditional adversarial training~\citep{madry2017towards}). Hence, we first perturb the training points using the training loss gradient, and then maximize predictive entropy on these adversarially perturbed examples.

\begin{figure}[t]
  \centering
  \begin{subfigure}[b]{0.5\textwidth}
    \centering
    \includegraphics[width=\linewidth]{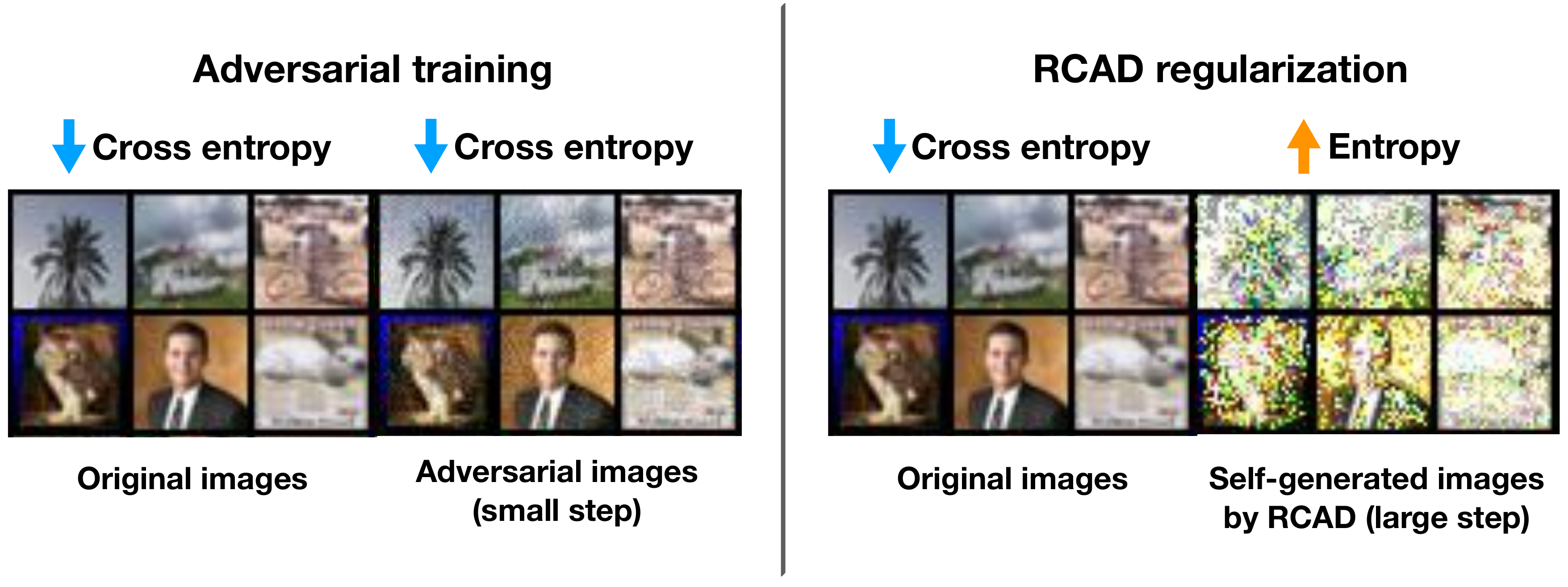}
    \caption{\label{fig:intro-right}}
  \end{subfigure}\hspace{3em}
 \begin{subfigure}[b]{0.3\textwidth}
    \centering
    \includegraphics[width=\linewidth]{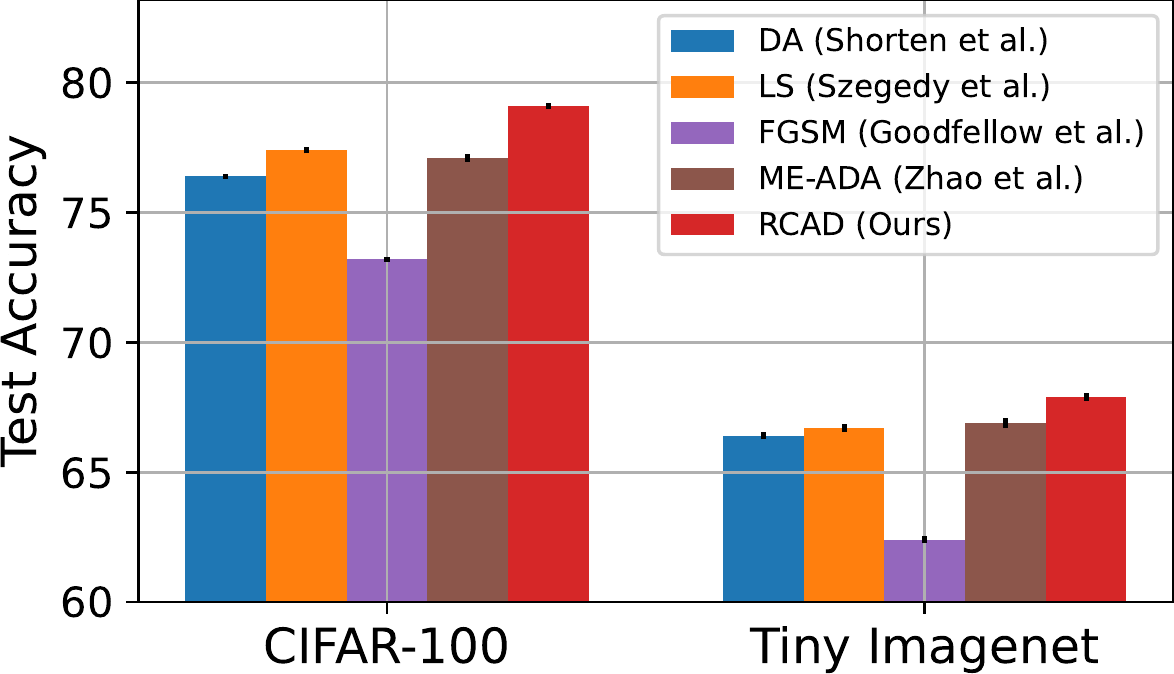}
    \caption{\label{fig:intro-center}}
  \end{subfigure}
 \caption{
 \footnotesize \textbf{Reducing confidence along adversarial directions} (RCAD) is a simple and  efficient regularization technique to improve test performance. 
 \emph{(Left)} For RCAD,  examples are generated by taking a large step ($10\times$ typical for adversarial examples) along the gradient direction. We see that generated images thus look very different from the original, with accentuated spurious components responsible for the model's flipped predictions on adversarial images.
 \emph{(Right)} RCAD achieves greater test accuracy than data augmentation (DA), label smoothing (LS), and methods that minimize cross-entropy on adversarial examples: adversarial training via  FGSM~\citep{goodfellow2014explaining} and ME-ADA~\citep{zhao2020maximum}.
 }
 \label{fig:intro-figure}
\end{figure}

In contrast to adversarial training, our method does not try to robustify the model to output the original label, but rather regularizes it to have reduced confidence on examples generated via a large step size (Figure~\ref{fig:intro-right}). As shown in Figure~\ref{fig:intro-center}, this can lead to significant improvements in in-distribution test accuracy unlike adversarial training~\citep{madry2017towards,miyato2018virtual,carlini2017towards,carlini2019evaluating}, which tends to decrease in-distribution test performance~\citep{raghunathan2019adversarial,zhang2019theoretically,Tsipras2019RobustnessMB}. Compared with semi-supervised learning methods~\citep{grandvalet2004semi,wang2020tent}, our method does not require an additional unlabeled dataset (it generates one automatically), and different from prior works~\cite{wang2020tent,zhou2021training} it also trains the model to be less confident on these examples.

As the {self-generated} samples in our procedure no longer resemble in-distribution data, we are uncertain of their labels and train the model to predict a uniform distribution over labels on them, following the principle of maximum entropy~\cite{jaynes1957information}. A few prior works~\cite{szegedy2016rethinking,pereyra2017regularizing,saito2019semi,dubey2018maximum} have also considered using the same principle to prevent the memorization of training examples, but only increase entropy on \iid sampled \emph{in-distribution} labeled/unlabeled data. In contrast, we study the effect of maximizing entropy on self-generated \emph{out-of-distribution} examples along adversarial directions.

The main contribution of this work is a training procedure we call \textit{RCAD: Reducing Confidence along Adversarial Directions}. RCAD improves test accuracy (for classification) and log-likelihood (for regression) across multiple supervised learning benchmarks. 
Importantly, we find that the benefits of RCAD are complementary to prior methods: Combining RCAD with alternative regularizers (e.g.,  augmentation~\citep{shorten2019survey,yun2019cutmix}, label smoothing~\citep{szegedy2016rethinking}, MixUp training~\citep{zhang2018mixup}) further improves performance. 
Our method requires adding $\sim$5 lines of code and is computationally efficient (with training time at most $1.3\times$ standard). 
We provide a theoretical analysis that helps to explain RCAD's benefits in a simplified setting, showing that RCAD can {unlearn} spurious features in the training data, thereby improving accuracy on unseen examples.

\section{Related Work}
\label{sec:prior-work}

\begin{table*}[t]
    \renewcommand{\arraystretch}{1.45}
    \caption{\footnotesize\textbf{Regularization objectives}: We summarize prior works that employ adversarial examples or directly regularize model's predictions $p_\vw(\ry\mid\rvx)$ along with the scalar hyperparameters (in $[\cdot]$) associated with each.}
    \label{tab:objectives}
    \footnotesize
    \centering
    \begin{tabular}{cl} 
        name  & objective \\ \hline
        cross entropy & $\min_{\vw} -\sum_{{\rvx, \ry} \in \hD} \log \pw(\ry \mid \rvx)$ \\ \hline
         label smoothing $[\epsilon]$~\citep{muller2019does}   & \multirow{1}{*}{$\min_\w -\sum_{{\rvx, \ry} \in \hD} \paren{(1 - \epsilon) \log \pw(\ry \mid \x) + \sum_{y' \neq \ry} \frac{\epsilon}{|\gY| - 1} \log \pw(y' \mid \x)}$} \\ \hline
        Adv. training $[\alpha]$~\citep{madry2017towards} & \multirow{1}{*}{$\min_\vw -\sum_{{\rvx, \ry} \in \hD} \log \pw(\ry \mid \rvx - \alpha\cdot \mathrm{sign} (\nabla_\rvx  \log \pw(\ry \mid \rvx)))$} \\ \hline
        \multirow{2}{*}{ME-ADA $[\alpha, \beta]$~\citep{zhao2020maximum}} & 
            $ \min_\vw -\sum_{(\rvx, \ry) \in \hD \cup \hat{\calD}'} \log \pw(\ry \mid \rvx) \quad \textrm{where, for a distance metric } C_\vw\mathrm{:} (\gX \times \gY) \times (\gX \times \gY) \mapsto \Real$ \\ 
         & $\hat{\calD}' \triangleq \{(\tilde{\vx}, \ry) \mid  \tilde{\vx} \triangleq \sup_{\vx_0} -\log p_w(\vx_0 \mid \ry) + \alpha \gH_\vw(\vx_0) - \beta C_\vw((\vx_0, \y), (\rvx, \ry)),\, \forall (\rvx, \ry) \in \hD \}$ \\ \hline
        RCAD (ours) $[\alpha, \lambda]$ & $\min_\vw \sum_{{\x, \y} \in \hat{\calD}} \paren{-\log \pw(\ry \mid \rvx) - \lambda \cdot \gH_\vw(\rvx - \alpha \cdot \nabla_\rvx \log \pw(\ry \mid \rvx))}$ \\ 
    \end{tabular}
\end{table*}

Below we survey common regularization techniques in machine learning, as well as other methods that utilize entropy maximization in training for various purposes (differing from our own).

\textbf{Data augmentation.} A common strategy to improve test accuracy (particularly on image tasks) is to augment the training data with corruptions~\citep{shorten2019survey} or surface level variations~\citep{devries2017improved, yun2019cutmix} (\eg rotations, random crops).
Some methods~\citep{miyato2018virtual, madry2017towards, goodfellow2014explaining} further augment data with imperceptibly different adversarial examples or interpolated samples~\citep{zhang2018mixup}.
Others~\citep{Poursaeed_2021_ICCV,odena2016semi} sample new examples from generative models that model the marginal density of the input distribution. 
Also related are semi-supervised methods, which  assume access to an extra set of unlabeled data and train the model to have more confident predictions on them~\citep{grandvalet2004semi, miyato2018virtual}.
All these methods \textit{minimize entropy}~\citep{grathwohl2019your} on the augmented or unlabeled data to improve generalization~\citep{zhang2018mixup, zhao2020maximum}. In contrast, our method \textit{maximizes entropy} on perturbed samples along the adversarial direction.

\textbf{Adversarial training.} 
Deep learning models are vulnerable to worst-case perturbations that can flip the model's predictions~\citep{goodfellow2014explaining,sabour2016adversarial}. Adversarial training was proposed to improve robustness to such attacks by reducing worst-case loss in small regions around the training samples~\citep{madry2017towards, kurakin2016adversarial, raghunathan2019adversarial}.
More recently, \citet{zhao2020maximum} proposed maximum entropy adversarial data augmentation (ME-ADA), which uses an information bottleneck principle to identify worst-case perturbations that both maximize training loss and predictive entropy. Similar to adversarial training ME-ADA still minimizes cross-entropy to output the same label on new points.  
There are two key differences between above methods and RCAD; \emph{(i)} rather than minimizing cross entropy loss on adversarial examples, we \emph{increase} model's uncertainty on the self-generated examples; and \emph{(ii)} we take much \emph{larger} steps---so 
 large that unlike adversarial images, the generated example is no longer similar to the original one (Figure~\ref{fig:intro-right}).
We show that these differences are important to improve test performance since adversarial training has been shown to hurt generalization on non-adversarial in-distribution test points \citep{raghunathan2019adversarial,tsipras2018robustness,zhang2019theoretically}. 
Further, RCAD has a lower (at most $1/5^{\textrm{th}}$) computational cost compared to multi-step adversarial training procedures~\citep{zhao2020maximum,zhang2019theoretically}.

\textbf{Robust objectives.} In order to improve robustness 
to noise and outliers in the data, a common approach is to modify the objective by considering risk averse loss functions~\citep[e.g.,][]{hertz2018introduction,rockafellar2000optimization,tamar2015optimizing} or incorporating regularization terms such as the $l_2/l_1$ norms~\citep{krogh1992simple,tibshirani1996regression}. 
Our method is similar to these approaches in that we propose a new regularization term. However, whereas most regularization terms are applied to the model's weights or activations, ours is directly applied to the model's predictions. Another effective loss function for classification problems is label smoothing~\citep{szegedy2016rethinking,church1991comparison,johnson1932probability},
which uniformly increases model's predictive uncertainty in a region around training samples~\citep{gao2020towards}. 
In contrast, RCAD increases entropy only on examples generated along the adversarial direction that has been shown to comprise of spurious features~\citep{ilyas2019adversarial, chen2020self}, thereby \emph{unlearning} them.

\textbf{Entropy maximization.} 
Finally, we note that our work builds upon prior work that draws connections between entropy maximization in supervised ~\citep{saito2019semi,dubey2018maximum,pereyra2017regularizing} and reinforcement learning~\citep{haarnoja2018soft}.
For example, \citet{pereyra2017regularizing} apply a hinge form of confidence penalty directly on training data which is similar to label smoothing in principle and performance (Table 2 in \cite{pereyra2017regularizing}). Other works like~\cite{dubey2018maximum,saito2019semi} also adapt the principle of entropy maximization but do so either on additional unlabeled data or a subset of the training samples. 
More recently \cite{pinto2021mix} show that maximizing entropy on interpolated samples from the same class improves out-of-distribution uncertainty quantification. 
In contrast to the above, we minimize cross-entropy loss on training data and only maximize entropy on samples generated along the adversarial direction. Our experimental results also span a wider set of benchmarks and presents significant gains $+1$--$3\%$ complementary to methods like label smoothing. We also theoretically analyze our objective for the class of linear predictors and show how RCAD can mitigate vulnerability to spurious correlations.

\section{RCAD: Reducing Confidence Along Adversarial Directions}
\label{sec:method}

We now introduce our regularization technique for reducing confidence along adversarial directions (RCAD). This section describes the objective and an algorithm for optimizing it; Section~\ref{sec:analysis} presents a more formal discussion of RCAD and in Section~\ref{sec:experiment} we provide our empirical study.

\textbf{Notation.} We are given a \emph{training} dataset $\hat{\calD} \triangleq \cbrck{(\rvx^{(i)}, \ry^{(i)})}_{i=1}^{N}$ where $\rvx^{(i)} \in \calX,\, \ry^{i} \in \gY$, are sampled \iid from a joint distribution $\calD$ over $\calX \times \calY$. We use $\hat{\calD}$ to denote both the training dataset and an empirical measure over it. The aim is to learn a parameterized distribution $p_\vw(\ry \mid \rvx),\; \vw \in \gW$ where the learnt model is typically obtained by maximizing the log-likelihood over $\hat \gD$. Such a solution is referred to as the maximum likelihood estimate (MLE):
$\hat{\vw}_{\mathrm{mle}}  \triangleq \argmax_{\vw \in \calW} \; \E_{\hat{\calD}}  \log p_\vw(\ry \mid \rvx)$.
In the classification setting, we measure the performance of any learned solution $\hat{\vw}$ using its accuracy on the \emph{test} (population) data: $\E_{\calD}\left[ \I(\argmax_{y'} p_{\hat{\vw}}(y' \mid \rvx) = \ry)\right]$.

Solely optimizing the log-likelihood over the training data can lead to poor test accuracy, since the estimate $\hat{\vw}_{\mathrm{mle}}$ can overfit on noise in $\hat{\gD}$ and fail to generalize to unseen samples. Many algorithms aim to mitigate overfitting by either designing suitable loss functions replacing the log-likelihood objective, or by augmenting the training set with additional data (see Section~\ref{sec:prior-work}). 
Our main contribution is a new data-dependent regularization term
that will depend not just on the model parameters but also on the training dataset. We reduce confidence on out-of-distribution samples that are obtained by perturbing the original inputs along the direction that adversarially maximizes training loss. 

In the paragraphs that follow we describe the methodology and rationale behind our objective that uses the following definition of model's predictive entropy when $\gY$ is a discrete set: 
\begin{align*}
    \calH_\vw(x) \; \triangleq \; - \sum_{y \in \gY} p_\vw(y \mid \rvx) \; \log p_\vw(y \mid \rvx).
\end{align*} 
In cases where $\gY$ is continuous, for example in regression tasks, we will use the  differential form of predictive entropy: $ - \int_{\gY} p_\vw(y \mid \rvx) \; \log p_\vw(y \mid \rvx) \; dy$.

\begin{wrapfigure}[8]{R}{0.4\textwidth}
\vspace{-2.4em}
\centering
\begin{minipage}{0.4\textwidth}
\begin{algorithm}[H]
\label{alg:mpe-algo}
\caption*{\footnotesize \textbf{RCAD: Reducing Confidence along Adversarial Directions}}
\begin{minted}[%
              mathescape=true,
              escapeinside=||,
              numbersep=3pt,
              gobble=0,
              fontsize=\footnotesize]{python}
def rcad_loss(x, y, |$\alpha$|, |$\lambda$|):
  loss = |$-$| model(x).log_prob(y)
  x_adv = x + |$\alpha$| * loss.grad(x)
  entropy = model(x_adv).entropy()
  return loss - |$\lambda$| * entropy
\end{minted}
\end{algorithm}
\end{minipage}
\end{wrapfigure}

\textbf{Reducing Confidence Along Adversarial Directions.} The key idea behind RCAD is that models should not only make accurate predictions on the sampled data, but also make uncertain predictions on examples that are very different from training data.
We use directions that adversarially maximize the training loss locally around the training points to construct these out-of-distribution examples that are different from the training data. This is mainly because adversarial directions have been known to comprise of spurious features~\cite{ilyas2019adversarial} and we want to regularize the model in a way that makes it uncertain on these features.  
We first describe how these examples are generated, and then describe how we train the model to be less confident. 

We generate an out-of-distribution example $\tilde{\rvx}$ by taking a large gradient of the MLE objective with respect to the input $\rvx$, using a step size of $\alpha > 0$:

\begin{align}
\tilde{\rvx} \; \triangleq \; \rvx - \alpha \cdot \nabla_{\rvx} \log p_\vw(\ry \mid \rvx)
\label{eq:self-gen-example}    
\end{align}

We train the model to make unconfident predictions on these self-generated examples by maximizing the model's predictive entropy. We add this entropy term $\calH_\vw\paren{\tilde{\rvx}}$ to the standard MLE objective, weighted by scalar $\lambda > 0$, yielding the final RCAD objective:

\begin{align}
    \hat{\vw}_{\textrm{rcad}} \;\; &\triangleq 
    \;\; \argmax_{\vw \in \calW} \;   \E_{\hat{\calD}} \brck{ \log p_\vw(\ry \mid \rvx) + \lambda \cdot \calH_\vw\paren{\rvx - \alpha \cdot \nabla_{\rvx} \log p_\vw(\ry \mid \rvx)} } 
    \label{eq:mpe-main-eq}
\end{align}

In adversarial training, adversarial examples are generated by solving a constrained optimization problem~\cite{madry2017towards,raghunathan2019adversarial}. Using a first-order approximation, the adversarial example $\tilde{\rvx}$ generated by one of the simplest solvers~\cite{goodfellow2014explaining} has a closed form resembling \eqref{eq:self-gen-example}. We note that
RCAD is different from the above form of adversarial training in two ways. First, RCAD uses a much larger step size ($10\times$ larger), so that the resulting example no longer resembles the training examples (Figure~\ref{fig:intro-right}). Second, whereas adversarial training updates the model to be more confident on the new example, RCAD trains the model to be less confident. Our experiments in Section~\ref{sec:experiment} (Figure~\ref{fig:addnl-results}b) show that these differences are important for improving test accuracy.

\textbf{Informal understanding of RCAD.} For image classification, image features that are pure noise and independent of the true label can still be spuriously correlated with the labels for a finite set of \iid drawn examples~\cite{singla2021salient}. Such features are usually termed \emph{spurious features}~\cite{zhang2021adversarial}. Overparameterized neural networks have a tendency to overfit on spurious features in the training examples~\citep{zhang2021understanding,sagawa2020investigation}. In order to generate unrealistic out-of-distribution samples, we pick examples that are far away from the true data point along the adversarial direction because \emph{(i)} adversarial directions are comprised of noisy features that are spuriously correlated with the label on a few samples~\cite{ilyas2019adversarial}; and \emph{(ii)} maximizing entropy on samples with an amplified feature forces the model to quickly unlearn that feature. By using a large step size $\alpha$ we exacerbate the spurious features in our self-generated example $\tilde{\rvx}$. When we maximize predictive entropy over our generated examples we then force the model to \emph{unlearn} these spurious correlations. Hence, the trained model can generalize better and achieve higher prediction performance on unseen test examples. In Section~\ref{sec:analysis} we build on this informal understanding to present a more formal analysis on the benefits of RCAD.

\section{Experiments: Does RCAD Improve Test Performance?}
\label{sec:experiment}
Our experiments aim to study the effect that RCAD has on test accuracy, both in comparison to and in addition to existing regularization techniques and adversarial methods, so as to understand the degree to which its effect is \emph{complementary} to existing methods. 

\textbf{Benchmark datasets.} We use six image classification benchmarks. In addition to CIFAR-10, CIFAR-100 \cite{krizhevsky2009learning}, SVHN \cite{netzer2011reading} and Tiny Imagenet \cite{Le2015TinyIV}, we modify CIFAR-100  by randomly sub-sampling 2,000 and 10,000 training examples (from the original 50,000) to create CIFAR-100-2k and CIFAR-100-10k. These smaller datasets allow us to study low-data settings where we expect the generalization gap to be larger. CIFAR-100(-2k/10k) share the same test sets. If the validation split is not provided by the benchmark, we hold out $10\%$ of our training examples for validation.

\textbf{Implementation details\footnote{Code for this work can be found at \url{https://github.com/ars22/RCAD-regularizer}.}.} Unless specified otherwise, we train all methods using the ResNet-18~\cite{he2016deep} backbone, and to accelerate training loss convergence we clip gradients in the $l_2$ norm (at $1.0$)~\cite{zhang2019gradient,hardt2016train}. 
We train all models for $200$ epochs and use SGD with an initial learning rate of $0.1$ and Nesterov momentum of $0.9$, and decay the learning rate by a factor of $0.1$ at epochs $100, 150$ and $180$~\cite{devries2017improved}. We select the model checkpoint corresponding to the epoch with the best accuracy on validation samples as the final model representing a given training method. For all datasets (except CIFAR-100-2k and CIFAR-100-10k for which we used $32$ and $64$ respectively) the methods were trained with a batch size of $128$. For details on algorithm specific hyperparameter choices refer to Appendix~\ref{app:sec:additional-exp-results}.

\textbf{Baselines.} The primary aim of our experiments is to study whether entropy maximization along the adversarial direction shrinks the generalization gap. Hence, we  explore baselines commonplace in deep learning that either \emph{(i)} directly constrain the model's predictive distribution on observed samples (label smoothing~\citep{muller2019does}) or \emph{(ii)} implicitly regularize the model by training on additional images generated in an adversarial manner. For the first, our main comparisons are to label smoothing~\cite{szegedy2016rethinking}), standard data augmentation~\cite{shorten2019survey}, cutout data augmentation~\cite{yun2019cutmix,devries2017improved}, and MixUp~\cite{zhang2018mixup} training. 
For the second, we compare with adversarial training\cite{madry2017towards} that uses FGSM~\cite{goodfellow2014explaining} to perturb the inputs. %
Additionally, we compare RCAD with two recent approaches that use adversarial examples in different ways: adversarial data augmentation (ADA) \cite{volpi2018generalizing} and maximum entropy adversarial data augmentation (ME-ADA) \cite{zhao2020maximum}. We summarize these baselines in Table~\ref{tab:objectives}.
\begin{figure*}[t]
  \centering
  \includegraphics[width=0.95\linewidth]{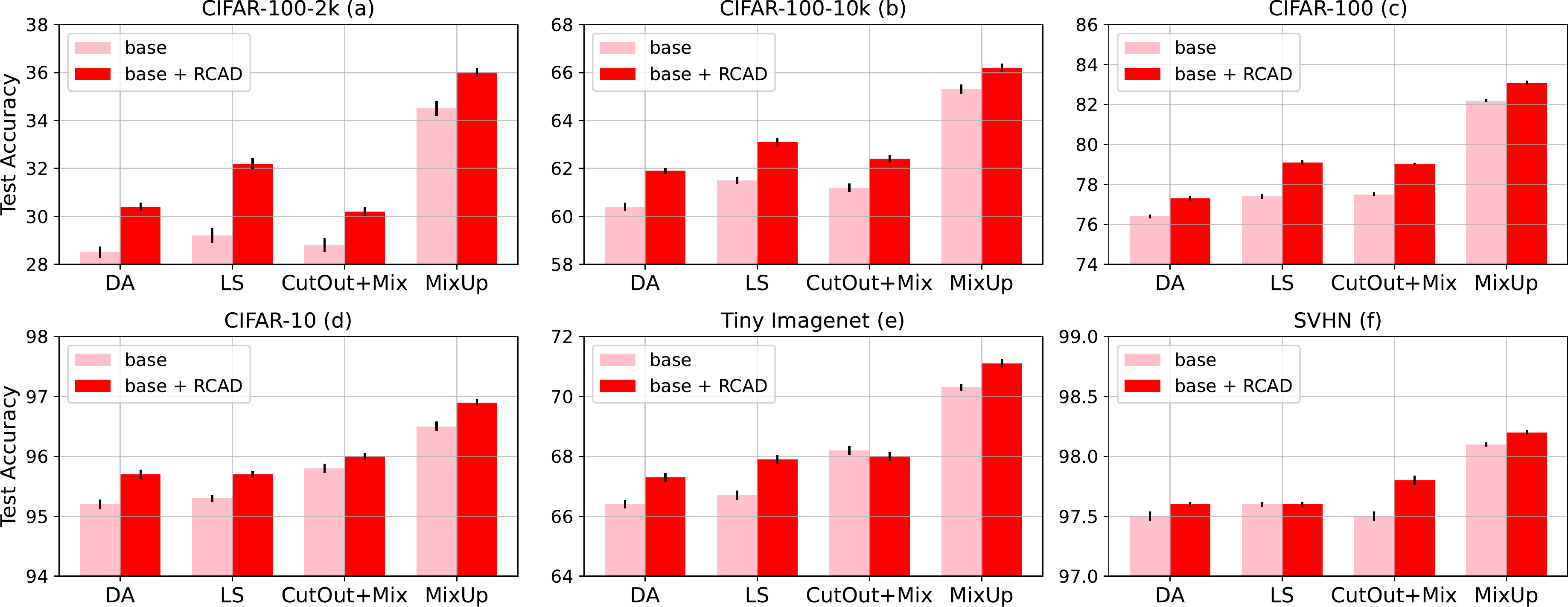}
 \caption{\footnotesize \textbf{Main results on supervised image classification benchmarks:} We plot the mean test accuracy and $95\%$ confidence intervals over $10$ independent runs for models trained with base methods: Data Augmentation (DA), Label Smoothing (LS), CutOut/CutMix (CutOut+Mix) augmentation, MixUp and compare them with the test accuracies of the models trained with the RCAD objective in \eqref{eq:mpe-main-eq}, in addition to the base methods.
 }
 \label{fig:main-results-base-ours}
\end{figure*}

\subsection{How much does RCAD improve test accuracy in comparison and in addition to other regularization methods?}
\label{subsec:MPE-test-acc}

Figure~\ref{fig:main-results-base-ours} presents the main empirical findings of RCAD over six image classification benchmarks and across four baseline regularizers. 
Across all datasets, we observe that training with the RCAD objective improves the test accuracy over the baseline method in $22/24$ cases. The effects are more pronounced on datasets with fewer training samples. For example, on CIFAR-100-2k, adding RCAD on top of label smoothing boosts the performance by $\approx 3\%$.
These results also show that the benefits of RCAD are complementary to prior methods---RCAD + MixUp outperforms MixUp, and RCAD + augmentation/smoothing outperforms both.
We test for statistical significance using a 1-sided $p$-value, finding that $p \ll 1e{-3}$ in $22 / 24$ comparisons.
In summary, these results show that our proposed regularizer is complementary to prior regularization techniques. 
\begin{figure}[t]
    \centering
    \hspace{2mm}
    \begin{subfigure}[b]{0.54\linewidth}
        \centering
        \includegraphics[width=\linewidth]{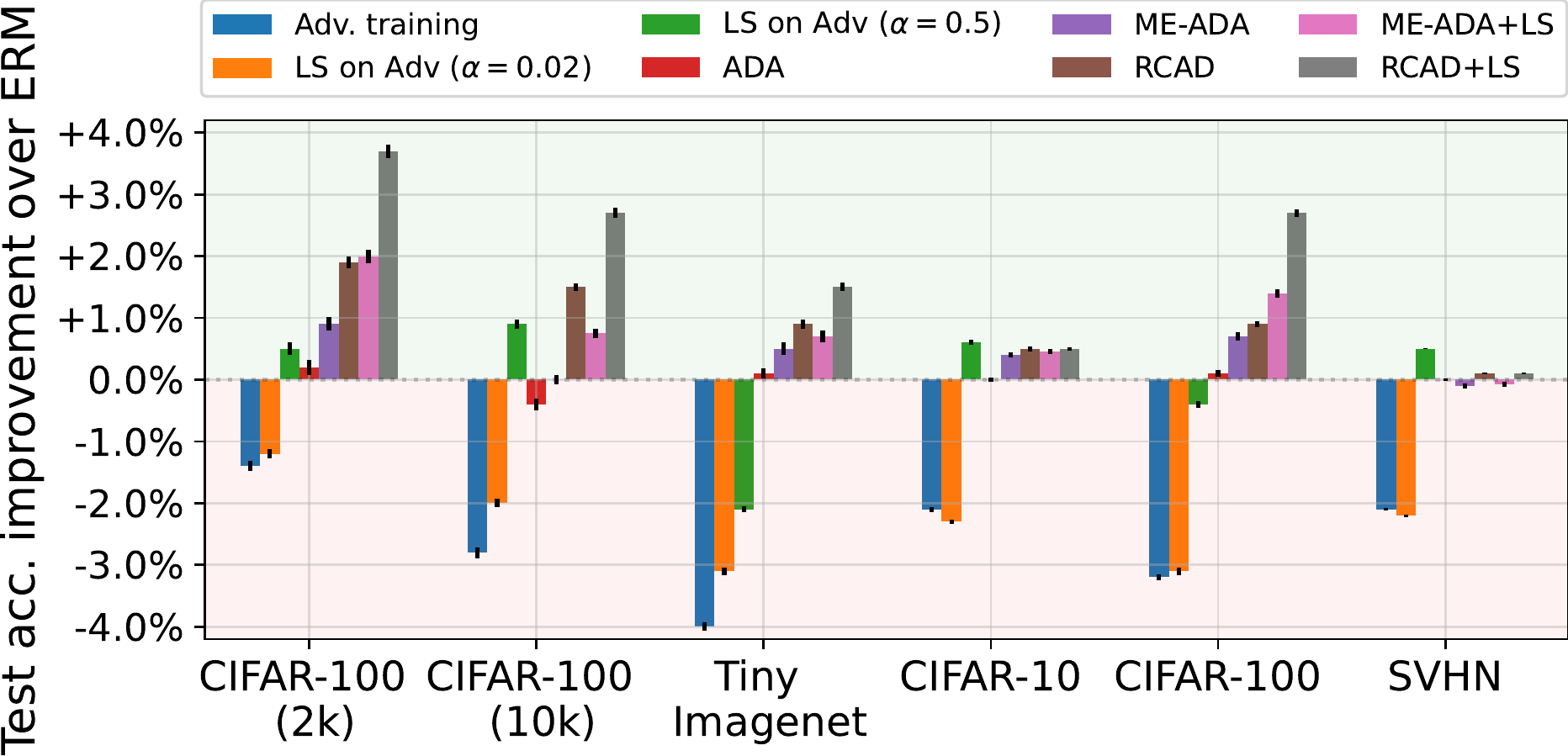}
    \caption{\label{fig:compare-left}}
    \end{subfigure}\hspace{0.52em}
    \begin{subfigure}[b]{0.42\linewidth}
        \centering
        \includegraphics[width=\linewidth]{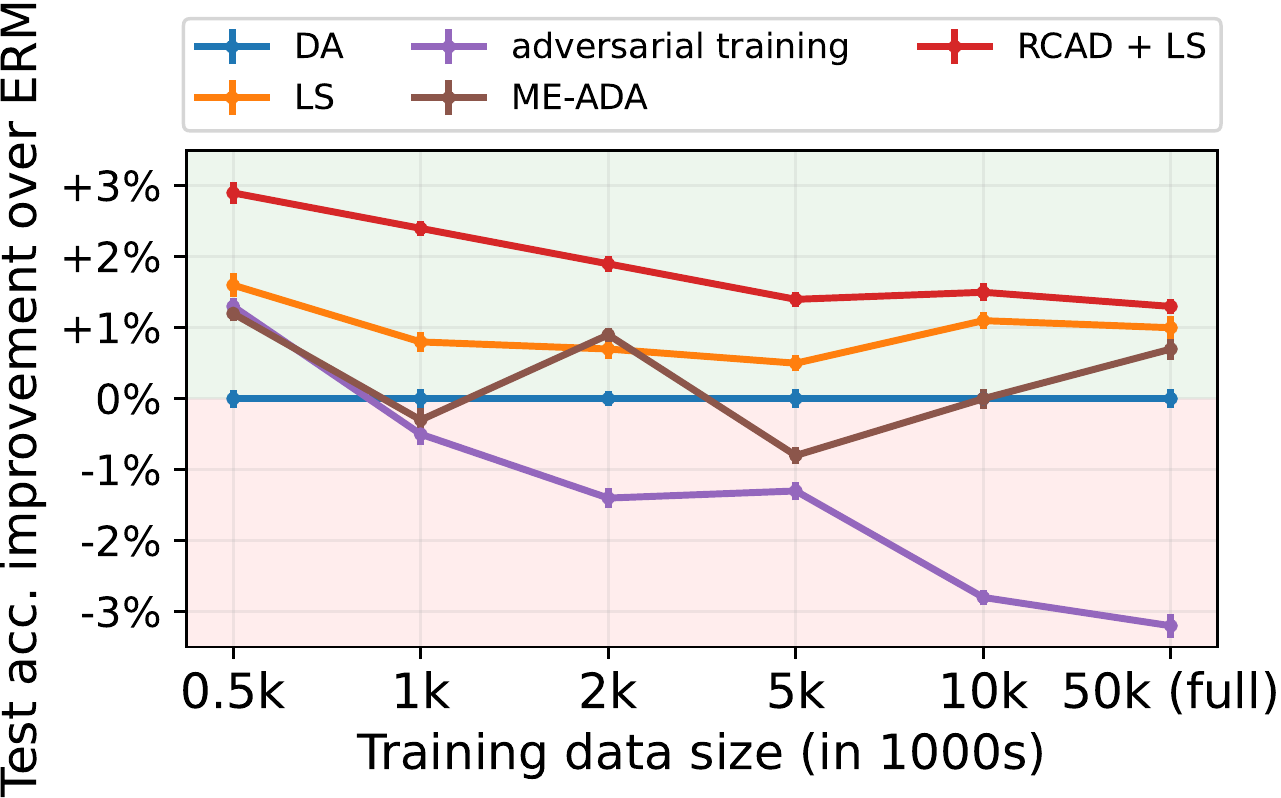}
    \caption{\label{fig:compare-right}}
    \end{subfigure}
    \caption{\footnotesize \figleft\,\textbf{How should we use adversarial examples to improve test accuracy?} We compare RCAD with adversarial baselines including label smoothing on adversarial samples (LS on Adv), measuring the improvement in test accuracy relative to ERM. We also compare RCAD with ME-ADA when combined with label smoothing. \figright\, \textbf{RCAD is more effective in low-data regime.} We compare the test accuracy improvement over ERM for baselines: Label Smoothing (LS), Data Augmentation (DA), adversarial training, and ME-ADA with RCAD + LS on different sub-samples of CIFAR-100 training set ($0.5$k $\rightarrow 50$k). We find RCAD achieves the largest gains in the low data regime. In both plots, we plot the mean and $95\%$ confidence intervals over $10$ independent runs. 
    \vspace{-0.1in}
    }
    \label{fig:compare-with-adv-and-low-data}
\end{figure}

\subsection{How effectively does RCAD improve test accuracy compared to adversarial training?}
\label{subsec:mpe-comparison-with-adv}
Our next set of experiments compares different ways of using adversarial examples. RCAD maximizes the model's predictive entropy on unlabeled examples along the adversarial direction (obtained with a large step-size $\approx0.5$). In contrast, other methods we compare against (Adversarial Training~\citep{madry2017towards}, ADA~\citep{volpi2018generalizing}, ME-ADA~\citep{volpi2018generalizing}) minimize the cross entropy loss on examples obtained by adversarially perturbing the inputs without changing the original labels (using a much smaller step size $\approx0.05$). Additionally, we look at the baseline that performs label smoothing on these adversarial samples (LS on Adv) -- a relaxed version of RCAD.
We evaluate all methods by measuring their test accuracy improvement over empirical risk minimization (ERM), noting that some of these methods were proposed to optimize robustness, a different metric. We show results in Figure~\ref{fig:compare-left} and note that RCAD outperforms adversarial training and LS on Adv with small step-sizes $(\alpha=0.02)$, ADA and ME-ADA by significant margins on all benchmarks. We numerically verify that RCAD statistically outperforms the best baseline ME-ADA by computing $1$-sided $p$-values ($p $=$0.009$ on CIFAR-10, $p<1e{-4}$ on others).  

Next, we look at LS on Adv with large $\alpha=0.5$. Since this is the same value of $\alpha$ used by RCAD for most datasets, this method is equivalent to performing label smoothing on examples generated by RCAD. Since label smoothing is a relaxed form of entropy maximization it is not surprising that the performance of this method is similar to (if not better than) RCAD on a few benchmarks.  
Finally, when both RCAD and ME-ADA are equipped with label smoothing, both methods improve, but the benefit of RCAD persists ($p$ $=$ $0.0478$ on CIFAR-10, $p<1e{-4}$ on others).

\subsection{How effective is RCAD in the low training data regime?}
\label{subsec:MPE-low-data}

Since supervised learning methods are more susceptible to overfitting when training samples are limited~\cite{goodfellow2016deep}, we analyze the effectiveness of RCAD in the low-data regime. To verify this, we sample small training datasets of size 0.5k, 1k, 2k, 5k and 10k from the 50,000 training samples in CIFAR-100. The test set for each is the same as that of CIFAR-100. We train the baseline regularizers and RCAD on each of these datasets and compare the observed improvements in test accuracies relative to ERM. We show our results for this experiment in Figure~\ref{fig:compare-right}.
Straight away we observe that RCAD yields positive improvements in test accuracy for any training dataset size. In contrast, in line with the \emph{robust overfitting} phenomena described in~\citet{raghunathan2019adversarial,raghunathan2020understanding,zhang2019theoretically}, adversarial training (with FGSM~\cite{goodfellow2014explaining}) is found to be hurtful in some settings.
Next we observe that compared to typical regularizers for supervised learning like label smoothing and data augmentation, the regularization effect of RCAD has a stronger impact as the size of the training data reduces. Notice that RCAD has a steeper upward trend (right$\rightarrow$left) compared to the next best method label smoothing, while outperforming each baseline in terms of absolute performance values.

\subsection{Additional Results and Ablations}
\label{subsec:addnl-results-ablations}
In addition to the experiments below on wider architectures, effect of step size $\alpha$, and regression tasks, we conduct more analysis and experiments for RCAD (\eg evaluating its robustness to  adversarial perturbations and distribution shifts). For these and other experimental details refer to Appendix~\ref{app:sec:additional-exp-results},~\ref{app:sec:mpe-robustness-to-shifts}.

\begin{figure*}
\hspace{0.5em}
\begin{minipage}[b]{0.23\textwidth}
    \includegraphics[width=\linewidth]{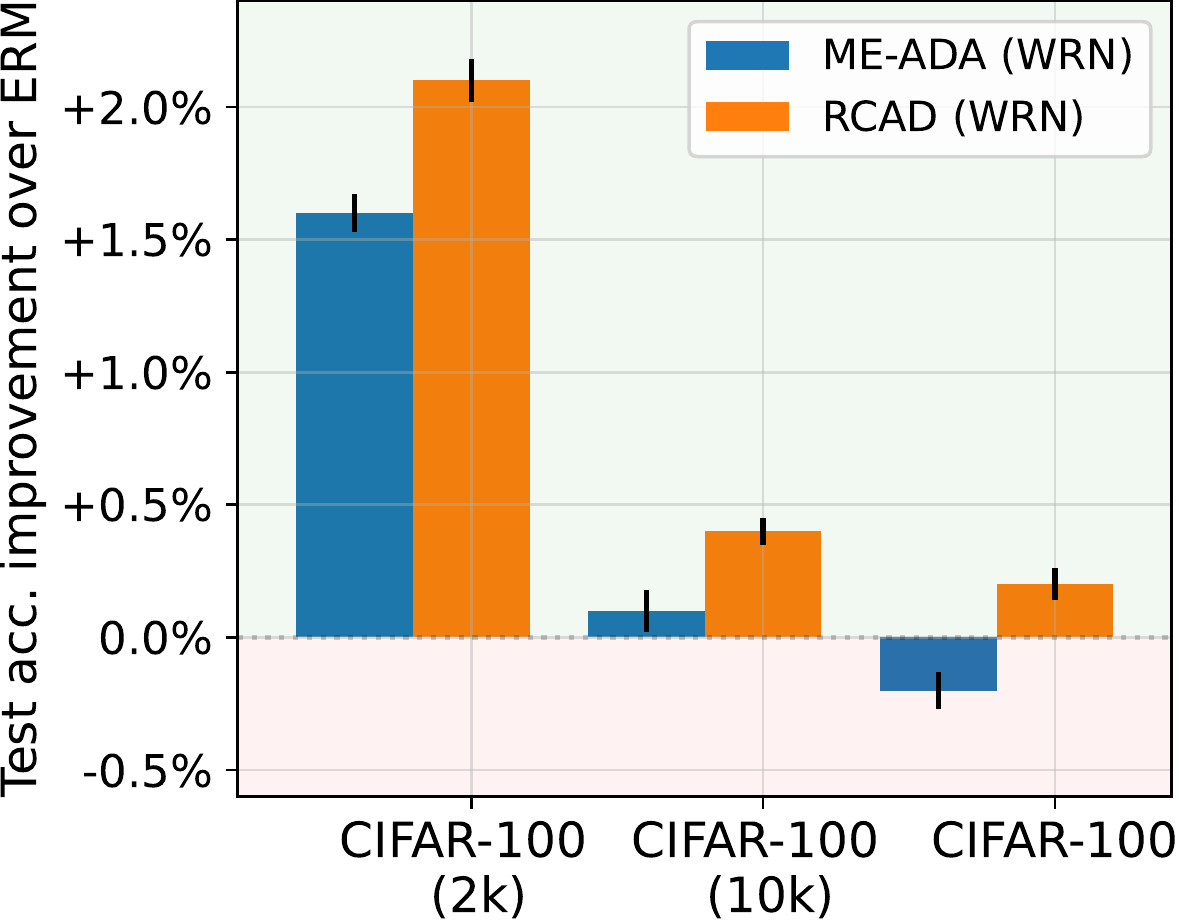}
    \caption*{(a)}
\end{minipage}\hspace{1em}
\begin{minipage}[b]{0.23\textwidth}
    \includegraphics[width=\linewidth]{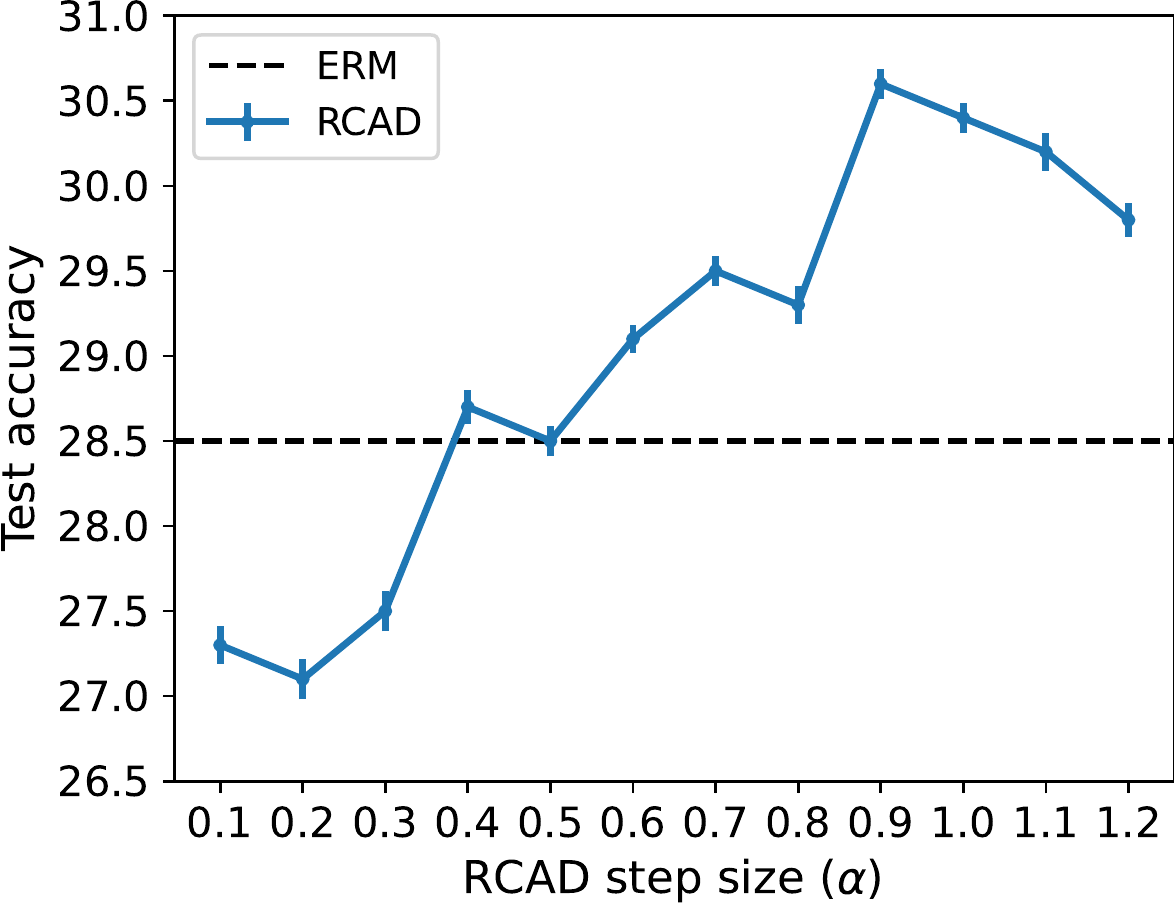}
    \caption*{(b)}
\end{minipage}\hspace{1em}
\begin{minipage}[b]{0.4\textwidth}
        \scriptsize
        \centering
        \setlength\tabcolsep{2pt}
        \renewcommand{\arraystretch}{1.2}
        \begin{tabular}{cccccc}\hline
    \multicolumn{6}{c}{Negative log-likelihood (NLL) on test data} \\ \hline
     Method & Boston & Concrete & Energy & Wine & Yacht \\ \hline
     ERM         & \vtop{{\hbox{\fontsize{9pt}{9pt} \strut $2.85$ }\hbox{\strut \fontsize{4pt}{4pt}$\pm0.06$}}} %
                 & \vtop{{\hbox{\fontsize{9pt}{9pt} \strut $3.25$ }\hbox{\strut \fontsize{4pt}{4pt}$\pm0.07$}}} %
                 & \vtop{{\hbox{\fontsize{9pt}{9pt} \strut $\mathbf{1.85}$ }\hbox{\strut \fontsize{4pt}{4pt}$\pm0.03$}}} %
                 & \vtop{{\hbox{\fontsize{9pt}{9pt} \strut $\mathbf{0.97}$ }\hbox{\strut \fontsize{4pt}{4pt}$\pm0.04$}}} %
                 & \vtop{{\hbox{\fontsize{9pt}{9pt} \strut $1.66$ }\hbox{\strut \fontsize{4pt}{4pt}$\pm0.04$}}} \\\hline %
     ME-ADA      & \vtop{{\hbox{\fontsize{9pt}{9pt} \strut $3.01$ }\hbox{\strut \fontsize{4pt}{4pt}$\pm0.04$}}} %
                 & \vtop{{\hbox{\fontsize{9pt}{9pt} \strut $\mathbf{3.14}$ }\hbox{\strut \fontsize{4pt}{4pt}$\pm0.03$}}} %
                 & \vtop{{\hbox{\fontsize{9pt}{9pt} \strut $1.99$ }\hbox{\strut \fontsize{4pt}{4pt}$\pm0.06$}}} %
                 & \vtop{{\hbox{\fontsize{9pt}{9pt} \strut $1.11$ }\hbox{\strut \fontsize{4pt}{4pt}$\pm0.05$}}} %
                 & \vtop{{\hbox{\fontsize{9pt}{9pt} \strut $1.78$ }\hbox{\strut \fontsize{4pt}{4pt}$\pm0.04$}}}  \\\hline %
     RCAD (ours)  & \vtop{{\hbox{\fontsize{9pt}{9pt} \strut $\mathbf{2.60}$ }\hbox{\strut \fontsize{4pt}{4pt}$\pm0.05$}}} %
                 & \vtop{{\hbox{\fontsize{9pt}{9pt} \strut $\mathbf{3.12}$ }\hbox{\strut \fontsize{4pt}{4pt}$\pm0.04$}}} %
                 & \vtop{{\hbox{\fontsize{9pt}{9pt} \strut $\mathbf{1.75}$ }\hbox{\strut \fontsize{4pt}{4pt}$\pm0.07$}}} %
                 & \vtop{{\hbox{\fontsize{9pt}{9pt} \strut $1.09$ }\hbox{\strut \fontsize{4pt}{4pt}$\pm0.05$}}} %
                 & \vtop{{\hbox{\fontsize{9pt}{9pt} \strut $\mathbf{1.54}$}\hbox{\strut \fontsize{4pt}{4pt}$\pm0.05$}}} \\ \hline %
    \end{tabular}
    \caption*{(c)}
\end{minipage}
\caption{\footnotesize \textbf{RCAD can be used with larger networks and is also effective on regression tasks.} 
\emph{(Left)}  We compare the improvements in test accuracy (over ERM) for RCAD and ME-ADA when all methods use Wide ResNet 28-10~\cite{zagoruyko2016wide}. 
\emph{(Center)} For CIFAR-100-2k, we show the effect on test accuracy of the step size $\alpha$ that RCAD takes while generating out-of-distribution examples along the adversarial direction. 
\emph{(Right)} We compare the negative log-likelihood on test samples for RCAD with baselines ERM and ME-ADA on five regression datasets from the UCI repository. For the above we show the mean and $95\%$ confidence intervals across $10$ runs. 
}
\label{fig:addnl-results}
\end{figure*}

\textbf{Results on a larger architecture.} We compare the test accuracies of RCAD and ME-ADA (most competitive baseline from ResNet-18 experiments) when trained with the larger backbone Wide ResNet 28-10~\cite{zagoruyko2016wide} (WRN) on CIFAR-100 and its derivatives. We plot these test accuracies relative to ERM trained with WRN in Figure~\ref{fig:addnl-results}a. Clearly, the benefit of RCAD over ME-ADA still persists albeit with slightly diminished absolute performance differences compared to ResNet-18 in Figure~\ref{fig:compare-left}. 

\textbf{Effect of step size $\alpha$ on test accuracy.} In Figure~\ref{fig:addnl-results}b we plot the test accuracy of RCAD on CIFAR-100-2k as we vary the step size $\alpha$. We see that RCAD is effective only when the step size is sufficiently large ($>0.5$), so that new examples do not resemble the original ones. When the step size is small $(<0.5)$, the perturbations are small and the new example is indistinguishable from the original, as in adversarial training. Our theory in Section~\ref{sec:analysis} also agrees with this observation. 

\textbf{Results on regression tasks.} On five regression datasets from the UCI repository we compare the performance of RCAD with ERM and ME-ADA in terms of test negative log-likelihood (NLL) ($\downarrow$ is better). To be able to compute NLL, we model the predictive distribution $p_{\vw}(\ry\mid \rvx)$ as a Gaussian and each method outputs two parameters (mean, variance of $p_{\vw}(\ry\mid \rvx)$) at each input $\rvx$. For the RCAD regularizer we use the differential form of entropy. Results are shown in Figure~\ref{fig:addnl-results}c from which we see that RCAD matches/improves over baselines on 4/5 regression datasets.

\section{Analysis: Why Does RCAD Improve Test Performance?}
\label{sec:analysis}

In this section, we present theoretical results that provide a more formal explanation as to why we see a boost in test accuracy when we regularize a classifier using RCAD. We analyze the simplified problem of learning $l_2$-regularized linear predictors in a fully specified binary classification setting. 
Specifically, our analysis will show that, while linear predictors trained with standard ERM can have a high dependence on spurious features, further optimization with the RCAD objective will cause the classifier to unlearn these spurious features.

The intuition behind our result can be summarized as follows: when RCAD generates examples after taking a large step along adversarial directions, a majority of the generated examples end up lying close to the decision boundary along the true feature, and yet a portion of them would significantly depend on the noisy spurious features. Thus, when RCAD maximizes the predictive entropy on these new examples, the classifier weights that align with the spurious components are driven toward smaller values.
Detailed proofs for the analysis that follows can be found in Appendix~\ref{app:sec:proofs-mpe-analysis}.

\paragraph{Setup.} We consider a binary classification problem with a joint distribution $\distr$
over $(\rvx, \ry) \in \gX \times \{0, 1\}$ where $\gX \subseteq \Real^{d+1}$ with $\rvx = \brck{\xone, \xtwo}^\top$, $\xone \in \Real,\; \xtwo \in \Real^d$ and for some  $\beta > 0$, $\distr$ is:
\begin{align}
    \label{eq:analysis-sample}
    \ry \sim \Unif \{-1, 1\}, \;\; \xone \sim \gN(\beta \cdot \ry, \sigma_1^2), \;\; \xtwo \sim \calN({\mathbf{0}}, \Sigma_2), \;\; \textrm{and let}\;\; \tSigma \triangleq \paren{\begin{array}{cc}
        \sigma_1^2 & \mathbf{0}  \\
        \mathbf{0} &  \Sigma_2
    \end{array}}.
\end{align}
We borrow this setup from~\citet{chen2020self} which analyzes self-training/entropy minimization on unlabeled out-of-distribution samples. While their objective is very different from RCAD, their setup is relevant for our analysis since it captures two kinds of features: $\xone$, which is the \emph{true} univariate feature that is predictive of the label $\ry$, and $\xtwo$, which is the high dimensional \emph{noise}, that is not correlated with the label under the true distribution $\gD$, but \whp is correlated with the label on finite sampled training dataset $\empdistr = \{(\rvx^{(i)}, \ry^{(i)})\}_{i=1}^{n} \sim \distr^n$. We wish to learn the class of homogeneous linear predictors $\gW \triangleq \{\vw \in \Real^{d+1}: \wtx = \wone\cdot\xone + \wtwo \cdot \xtwo,  \norm{\vw}_2 \leq 1\}$. 
For mathematical convenience we make three modifications to the RCAD objective in \eqref{eq:mpe-main-eq} that are common in works analyzing logistic classifiers and entropy based objectives ~\cite{soudry2018implicit,chen2020self}:
\emph{(i)} we minimize the margin loss $\marginlosswxy \triangleq \max(0,  \gamma - \ry \cdot \wtx)$ (instead of the negative log-likelihood) over $\gD'$ with some $\gamma > 0$;
\emph{(ii)} we consider the constrained optimization problem in \eqref{eq:mpe-constr-obj} as opposed to the Lagrangian form of RCAD in \eqref{eq:mpe-main-eq}~\cite{li1995zero}; and \emph{(iii)} we use Lemma~\ref{lem:ent-approx} and replace $\gH_\vw(\rvx + \alpha \cdot \nabla_{\rvx} \marginlosswxy)$ with  $\exp(-|\innerprod{\vw}{\rvx + \alpha \cdot \nabla_{\rvx} \marginlosswxy}|)$.  Note that the optimal $\vw^* \in \gW$ that has the best test accuracy and lowest population margin loss is given by $\vw^* \triangleq \brck{1, 0, \dots, 0}^{\top}$.

\begin{align}
& \max_{\vw}\;\; \gM_{\hat \gD}(\vw) \triangleq \E_{\hat{\gD}} \; \exp(-|\innerprod{\vw}{\rvx + \alpha \cdot \nabla_{\vx} \marginlosswxy}|)  \nonumber\\
& \;\,\textrm{s.t.}\;\;\;\; \E_\empdistr \; \I (\marginlosswxy > 0) \leq \rho/2,\;\;  \vw \in \gW \label{eq:mpe-constr-obj}
\end{align}

\begin{figure}[t]
    \centering
    \hspace{2mm}
    \begin{subfigure}[b]{0.41\linewidth}
        \centering
        \includegraphics[width=\linewidth]{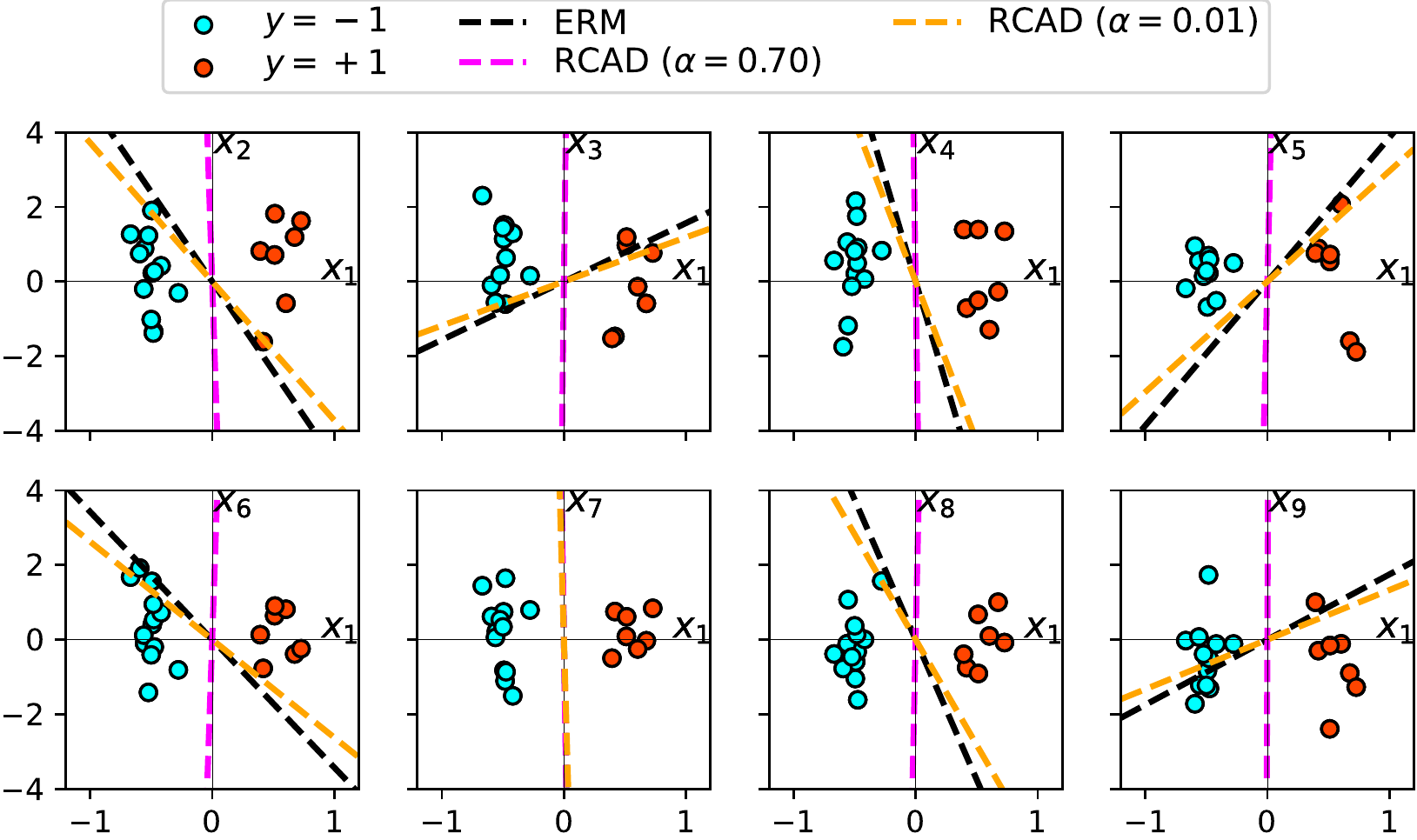}
    \caption{\label{fig:analysis-left}}
    \end{subfigure}
    \begin{subfigure}[b]{0.27\linewidth}
        \centering
        \includegraphics[width=\linewidth]{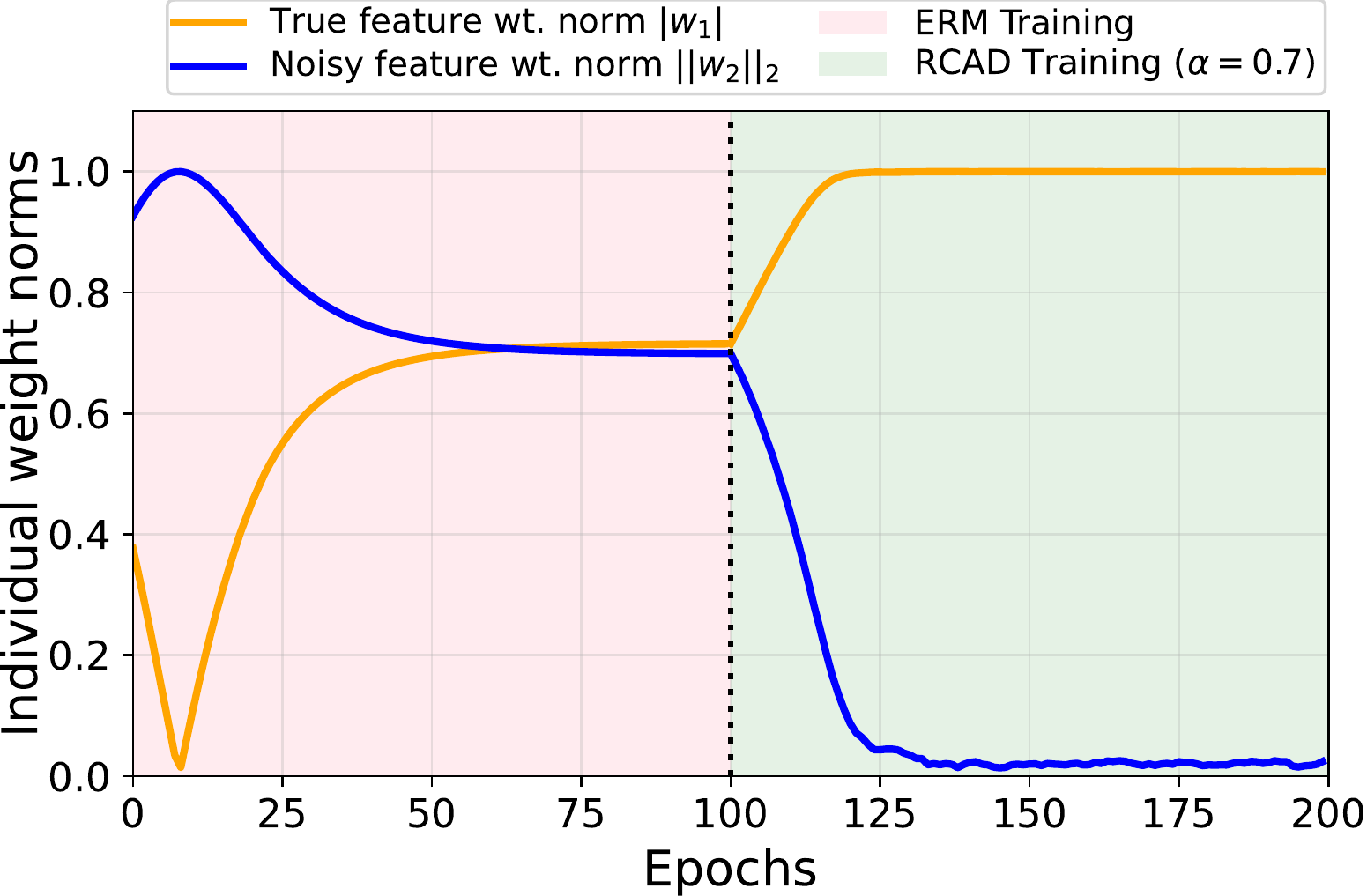}
    \caption{\label{fig:analysis-center}}
    \end{subfigure}
    \begin{subfigure}[b]{0.27\linewidth}
        \centering
        \includegraphics[width=\linewidth]{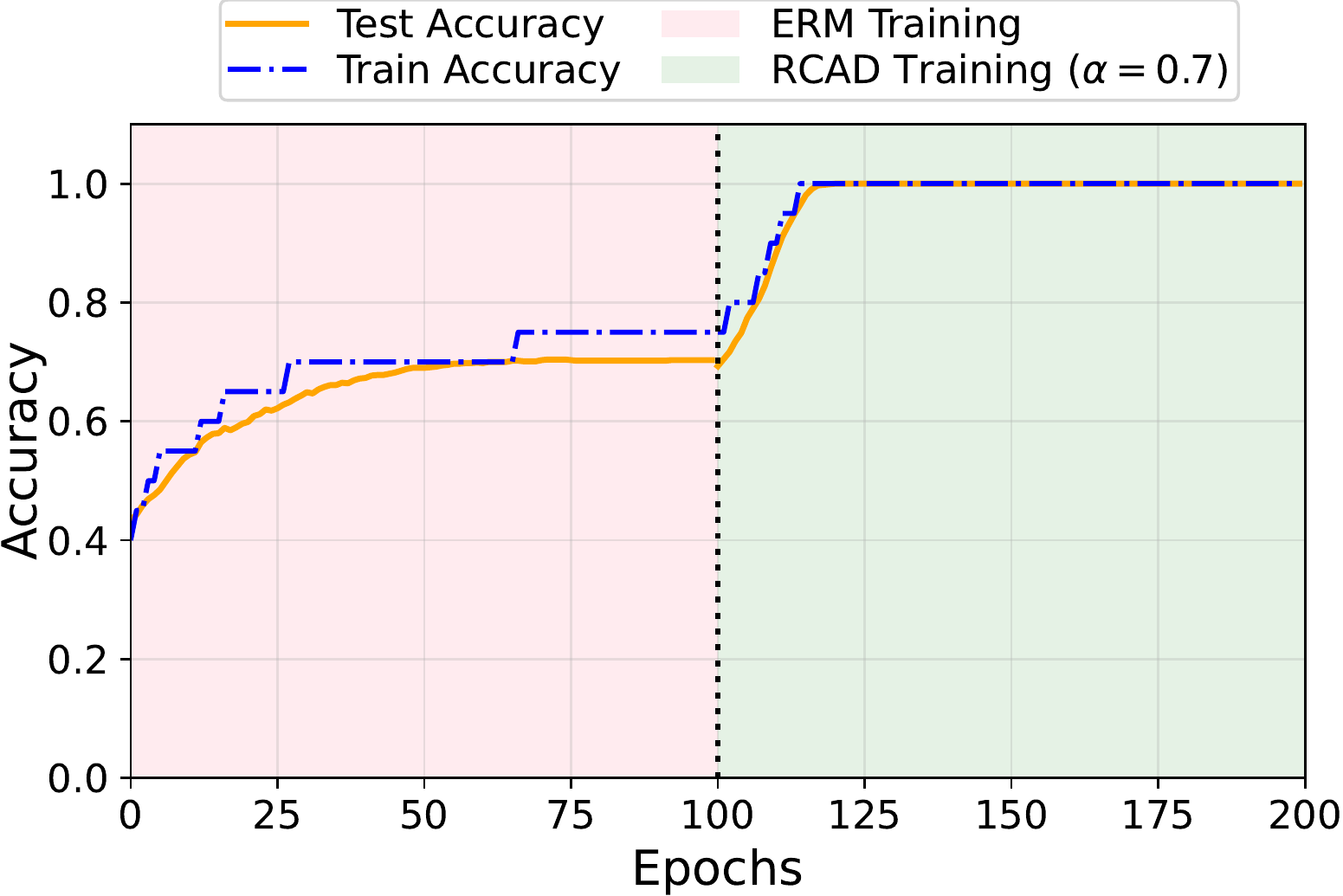}
    \caption{\label{fig:analysis-right}}
    \end{subfigure}
    \hfill
    \caption{\footnotesize \textbf{RCAD corrects ERM solution:}  
    \figleft\, We plot the training data and high-dimensional decision boundaries for all three estimates $\hat{\vw}_{\mathrm{erm}}$ and $\hat{\vw}_{\mathrm{rcad}}$ with $\alpha=0.01,0.70$. Here, we plot the linear boundary projected onto the set of planes $\{\xone, \ervx_i\}_{i=2}^{i=8}$ spanned by the true feature $\xone$ and noisy features $\ervx_2,\ldots,\ervx_8$. 
    Across the ERM and RCAD training iterations, we plot the weight norms along the true feature $|\wone|$ and noisy dimensions $\|\wtwo\|_2$ in the \emph{(Center)} plot, and on the \emph{(Right)} we plot the train/test accuracies.
    }
    \label{fig:analysis-figure}
\end{figure}

\begin{lemma}[\cite{chen2020robust,soudry2018implicit}, informal]
\label{lem:ent-approx}
 $\gH_\vw(\rvx)= \gH_{\mathrm{bin}}((1+\exp(-\innerprod{\vw}{\rvx}))^{-1})$ where $\gH_{\mathrm{bin}}(p)$ is the entropy of $\Bern(p)$ distribution. Thus
$\gH_\vw(\rvx) \approx \exp(-|\innerprod{\vw}{\rvx}|)$, as both exhibit same tail behavior. 
\end{lemma}
We analyze the RCAD estimate $\hat{\vw}_{\mathrm{rcad}} \triangleq \witer{T}$ returned after $T$ iterations of projected gradient ascent on the non-convex objective in \eqref{eq:mpe-constr-obj}, initialized with $\wzero$ satisfying the constraints in \eqref{eq:mpe-constr-obj}.  Given the projection $\Pi_\gS$ onto the convex set $\gS$, and learning rate $\eta > 0$, the update rule for $\witer{t}$ is as follows:
    $\tilde{\vw}^{(t+1)} =  \witer{t} + \eta \cdot \nabla_{\witer{t}} \gM_{\hat \gD}(\vw)$, and $\witer{t+1} = \Pi_{\|\vw\|_2 \leq 1} \tilde{\vw}^{(t+1)}$.
Since the initialization $\witer{0}$ satisfies the constraint in  \eqref{eq:mpe-constr-obj}, it has a low margin loss on training samples and using margin based generalization gaps~\cite{kakade2008complexity} we can conclude   
\whp $\wzero$ has low test error $(\leqslant \rho)$. This, in turn tells us that $\wzero$ should have learnt the true feature $x_1$ to some extent (Lemma~\ref{lem:init-cond}). 
Intuitively, if the classifier is not trained at all, then adversarial directions would be less meaningful since they may include a higher component of the true feature $\xone$, as opposed to noisy $\xtwo$. To obtain $\wzero$, we simply minimize loss $l_\gamma$ on $\hat \gD$ using projected (onto $\|\vw\|_2\leq 1$) gradient descent and the ERM solution $\hat{\vw}_{\mathrm{erm}}$ serves as the initialization $\witer{0}$ for RCAD's gradient ascent iterations. 
Note that $\hat{\vw}_{\mathrm{erm}}$ still significantly depends on spurious features (see Figure~\ref{fig:analysis-figure}). Furthermore, this dependence is unavoidable and worse when $n$ is small, or when the noise dimension $d$ is large.

\begin{lemma} [true feature is partially learnt before RCAD]
\label{lem:init-cond}
If $\wzero$ satisfies constraint in ~\eqref{eq:mpe-constr-obj}, and when $n \gsim
\frac{\beta^2 + \log (1/\delta) \cdot \|\tilde{\Sigma}\|_{\mathrm{op}}+\|\tSigma\|_*}{\gamma^2\rho^2}$,
with probability $1-\delta$ over 
$\hat{\gD}$, $\;\; \woneiter{0} \geq  \frac{\erfcinv{2\rho} \cdot \sqrt{2\sigma_{\mathrm{min}} (\tSigma)}}{\beta}$.
\end{lemma}
We are now ready to present our main results in Theorem~\ref{thm:main-result} which states that after $T = \gO(\log (1/\epsilon))$ gradient ascent iterations of optimizing the RCAD objective ${\hat{\vw}}_{\mathrm{rcad}}$ is $\epsilon$ close to $\vw^*$, since it \emph{unlearns} the spurious feature $(\|\wtwoiter{T}\|_2 \leq \epsilon)$ and \emph{amplifies} the true feature $(|\woneiter{T}| \geq \sqrt{1 - \epsilon^2})$. 
Note that Theorem~\ref{thm:main-result} suggests a minimum step size $\alpha$ in arriving at a sufficient condition for RCAD to unlearn spuriousness and improve test performance This is also in line with our empirical findings (Figure~\ref{fig:addnl-results}b). Since $\alpha = \Theta(\gamma)$, most of the points generated  by RCAD lie close to the decision boundary along $\ervx_1$, except for the ones with noisy features that are correlated with the classifier weights. 

\begin{theorem}[$\hat{\vw}_{\mathrm{rcad}} \rightarrow \vw^*$]
\label{thm:main-result}
If $\beta = \Omega(\|\tSigma\|_{\mathrm{op}})$ and $\exists\; c_0, K > 0$, such that
$\beta \gsim \alpha \gsim \max(\gamma, \|\tSigma\|_{\mathrm{op}})$ and $\gamma + c_0 \cdot \sigmaMintSigma \geq \beta$, then with $n \gsim \frac{\beta^2 + \log (1/\delta) \cdot \|\tilde{\Sigma}\|_{\mathrm{op}}+\|\tSigma\|_*}{\gamma^2\erfc{K\alpha/\sqrt{2\sigmaMintSigma}}^2} + \frac{\log{1/\delta}}{\epsilon^4}$, after $T = \gO\paren{\log(1/\epsilon)}$ gradient ascent iterations, $|\woneiter{T}| \geq \sqrt{1-\epsilon^2}$ and $\|{\wtwoiter{T}}\|_2 \leq \epsilon$, with probability $1-\delta$ over $\calD'$. 
\end{theorem}

The key idea behind our guarantee above is an  inductive argument. We show that, if the class separation $\beta$ and step size $\alpha$ are sufficiently larger than any noise variance, then RCAD monotonically increases the weight norm along the true feature and monotonically decreases it along the noisy dimensions with each gradient ascent update. This is because at any given instant the gradient of RCAD objective $\gM_\gD(\vw)$ with respect to $\wone$ always points in the same direction, while the one with respect to $\wtwo$ is sufficiently anti-correlated with the direction of $\wtwo$. We formalize this argument in Lemma~\ref{lem:update-cond}. It is also easy to see why the update will improve test accuracy monotonically, thus satisfying the constraint in \eqref{eq:mpe-constr-obj} with high probability.

\begin{lemma}[$\wone, \wtwo$ update]
\label{lem:update-cond}
If $\alpha, \beta, \gamma$ and sample size $n$ satisfy the noise conditions in Theorem~\ref{thm:main-result}, then $\langle \pardev{\gM_\gD(\vw)}{\woneiter{t}}, \woneiter{t} \rangle > 0$,
and $|\wtildeoneiter{t+1}| > |\woneiter{t}|$. On the other hand, $\exists c_1 > 0$ such that $\langle \nabla_{\wtwoiter{t}}{\gM_\gD(\witer{t})}, \wtwoiter{t} \rangle \leq -c_1\cdot \|\wtwoiter{t}\|_2^2$. Consequently $\exists \eta$, such that $\|\wtildetwoiter{t+1}\|_2 / \|\wtwoiter{t}\|_2 < 1$.
\end{lemma}

\textbf{Empirically validating our theoretical results in the toy setup.} Now, using the same setup as our theoretical study, we check if taking the ERM solution and updating it with RCAD truly helps in unlearning the spurious feature $\xtwo$ and amplifying the true feature $\xone$. With $d=8, \gamma=2.0, \beta=0.5, \sigma_1=0.1, \Sigma = \mathbf{I}_{d}$ we collect training dataset $\hat \gD$ of size $20$ according to the data distribution in \eqref{eq:analysis-sample}. First, we obtain the ERM solution $\hat{\vw}_{\mathrm{erm}}$ by optimizing the margin loss on $\hat \gD$ for $100$ projected gradient descent iterations. Then, with  $\hat{\vw}_{\mathrm{erm}}$ as initialization we optimize the RCAD objective in \eqref{eq:mpe-constr-obj} for $100$ additional projected gradient ascent iterations to obtain RCAD estimate $\hat{\vw}_{\mathrm{rcad}}$. For RCAD, we try both small $(\alpha=0.01)$ and large  $(\alpha=0.7)$ values of the step size. Using the results of this study in Figure~\ref{fig:analysis-figure} we shall now verify our main theoretical claims: 

(i) \textbf{RCAD unlearns spurious features learnt by ERM (Theorem~\ref{thm:main-result}):} RCAD training with large step size $\alpha=0.7$ corrects ERM's decision boundaries across all noisy dimensions \ie $\hat{\vw}_{\mathrm{rcad}}$ has almost no dependence on noisy dimensions $\{\ervx_i\}_{i=2}^{i=8}$ (Figure~\ref{fig:analysis-left}). On the other hand, the ERM solution has a significant non-zero component across all noisy dimensions (except $\ervx_7$). 

(ii) \textbf{RCAD is helpful only when step size is large enough (Theorem~\ref{thm:main-result}):} In Figure~\ref{fig:analysis-left} we see that RCAD estimate with small step size $\alpha=0.01$ has a higher dependence on spurious features compared to ERM. Thus optimizing RCAD with a very small $\alpha$ leads to an even poor solution than ERM. This is because, the perturbed data point still has a significant component of the true feature, which the model is forced to unlearn when RCAD tries to maximize entropy over the perturbed point. 

(iii) \textbf{The weight norms $|\wone|$ and $\|\wtwo\|_2$ change monotonically through RCAD iterations (Lemma~\ref{lem:update-cond}):} In Figure~\ref{fig:analysis-center} we see that the norm along the true feature $|\wone|$ increases monotonically through RCAD iterations, whereas $\|\wtwo\|_2$ decreases monotonically, until $|\wone|\approx 1$ and $\|\wtwo\|_2\approx 0$. Thus, RCAD successfully recovers optimal $\vw^*$. Based on this, we also see the test accuracy increase monotonically through RCAD iterations; reaching $100\%$ by the end of it (Figure~\ref{fig:analysis-right}). In contrast, since the ERM solution depends on spurious features, its accuracy does not improve beyond $70\%$ in the first $100$ iterations. We find that training ERM for more iterations only improves training accuracy.   

The analysis above explains why RCAD improves test performance in the linear setting and it is inline with our intuition of RCAD's ability to unlearn spurious features. To check if our intuition also generalizes to deeper networks we further study RCAD's behaviour in a non-linear setting using a different toy example in Appendix~\ref{app:sec:toy-non-linear}. Furthermore, while our analysis is restricted to linear classifiers, linear models can provide a rough proxy for neural network learning dynamics via the neural tangent kernel (NTK) approximation~\citep{jacot2018neural}. This strategy has been used in a number of prior works~\citep{nagarajan2019uniform,tancik2020fourier,cao2019generalization} and extending the analysis to the NTK setting may be possible in future work.

\vspace{-0.1in}
\section{Conclusion}
\label{sec:conclusion}
\vspace{-0.1in}
In this paper we propose RCAD, a regularization technique that maximizes a model's predictive entropy on out-of-distribution examples. These samples lie along the directions that adversarially maximize the loss locally around the training points. 
Our experiments on image classification benchmarks show that RCAD not only improves test accuracy by a significant margin, but that it can also seamlessly compose with prior regularization methods.
We find that RCAD is particularly effective when learning from limited data. Finally, we present analyses in a simplified setting to show that RCAD can help the model unlearn noisy features.
Some current limitations of our work are that RCAD slows training by $30\%$ and our theoretical analysis is limited to the linear case, which would be interesting directions to address/expand on in future work---particularly in light of the significant empirical benefits of RCAD for improving generalization.

{
\textbf{Acknowledgements.} The authors would like to thank Oscar Li, Ziyu Liu, Saurabh Garg at Carnegie Mellon University and members of the RAIL lab at UC Berkeley for helpful feedback and discussion. 
}

{\footnotesize
\bibliographystyle{apalike}
\bibliography{main}
}

\appendix
\clearpage
\section*{Appendix Outline}

\ref{app:sec:proofs-mpe-analysis} Proofs for Section~\ref{sec:analysis}.

\ref{app:sec:additional-exp-results} Additional Details for Experiments in Section~\ref{sec:experiment}.

\ref{app:sec:mpe-robustness-to-shifts} RCAD can Improve Robustness to Adversarial/Natural Distribution Shifts.

\ref{app:sec:toy-non-linear} Analyzing RCAD vs. ERM in a Non-linear Toy Classification Setup.

\section{Proofs for Section~\ref{sec:analysis}}
\label{app:sec:proofs-mpe-analysis}

In Section~\ref{sec:analysis} of the main paper we analyze the performance of our method RCAD and compare it to empirical risk minimization (ERM) in a simplified Binary classification setting. 
Our investigation reveals that one reason for the better test performance of RCAD could be its ability to unlearn noisy features that are spuriously correlated with the labels in the finite sampled training dataset.  
This is identified under some conditions over the data distribution (noise conditions) and specifically when the RCAD objective is optimized using iterations of projected gradient ascent initialized with a decent (better than random) classifier. This implies that the initial classifier (initialized with the ERM solution in our case) has learnt the true feature to some extent. The solution obtained at the end of projected gradient ascent iterations on the RCAD objective maximizing entropy along adversarial directions, is termed as the RCAD solution.
We also matched some of our theoretical results to trends observed empirically with respect to both the decision boundary learnt by RCAD vs. ERM classifier and the effect of the step size $\alpha$ on the performance of RCAD (Figure~\ref{fig:analysis-figure}). 

In this section we present proofs for our theoretical claims in Section~\ref{sec:analysis}. 
For the benefit of the reader, we begin by redefining our setup with some additional notation, followed by technical details on how the new examples are generated by RCAD in this setup. 
We then provide an informal proof outline for our main result in Theorem~\ref{thm:main-result}.

\paragraph{Setup.} We consider a binary classification problem with a joint distribution $\distr$
over $(\rvx, \ry) \in \gX \times \{0, 1\}$ where $\gX \subseteq \Real^{d+1}$ with $\rvx = \brck{\xone, \xtwo}^\top$, $\xone \in \Real,\; \xtwo \in \Real^d$ and for some  $\beta > 0$, $\distr$ is:

\begin{align}
    \label{eq:analysis-sample-app}
    \ry \sim \Unif \{-1, 1\}, \;\; \xone \sim \gN(\beta \cdot \ry, \sigma_1^2), \;\; \xtwo \sim \calN({\mathbf{0}}, \Sigma_2) \nonumber \\ 
    \tSigma \triangleq \paren{\begin{array}{cc}
        \sigma_1^2 & \mathbf{0}  \\
        \mathbf{0} &  \Sigma_2 \\ 
    \end{array}}, \quad \sigmaw  \triangleq \sqrt{\vw^T \tSigma \vw}, \quad \aw \triangleq \frac{\beta \wone - \gamma}{\sigmaw} 
\end{align}

This setup is relevant for our analysis since it captures two kinds of features: $\xone$, which is the \emph{true} univariate feature that is predictive of the label $\ry$, and $\xtwo$, which is the high dimensional \emph{noise}, that is not correlated with the label under the true distribution $\gD$, but \whp is correlated with the label on finite sampled training dataset $\empdistr = \{(\rvx^{(i)}, \ry^{(i)})\}_{i=1}^{n} \sim \distr^n$. We wish to learn the class of homogeneous linear predictors $\gW \triangleq \{\vw \in \Real^{d+1}: \wtx = \wone\cdot\xone + \wtwo \cdot \xtwo,  \norm{\vw}_2 \leq 1\}$. 
Following the modifications we make to the RCAD objective in Section~\ref{sec:analysis}, and using Lemma~\ref{lem:ent-approx} we arrive at the following constrained form of the RCAD objective where $\marginlosswxy \triangleq \max(0,  \gamma - \ry \cdot \wtx)$.

\begin{align}
& \max_{\vw}\;\; \gM_{\hat \gD}(\vw) \triangleq \E_{\hat{\gD}} \; \exp(-|\innerprod{\vw}{\rvx + \alpha \cdot \nabla_{\vx} \marginlosswxy}|)  \nonumber\\
& \;\,\textrm{s.t.}\;\;\;\; \E_\empdistr \; \I (\marginlosswxy > 0) \leq \rho/2,\;\;  \vw \in \gW \label{eq:mpe-constr-obj-app}
\end{align}

Note that the gradient $\nabla_{\rvx} \marginlosswxy$ maybe undefined at the margins, when $y\cdot \innerprod{\vw}{\rvx} = \gamma$. Hence, we rely on subgradients defined in \eqref{eq:subgrad}. Since, many subgradient directions exist for the margin points, for consistency, we stick with $\partial_{\rvx} \marginlosswxy = \{\bf{0}\}$ when $y\cdot \innerprod{\vw}{\rvx} = \gamma$. Note, that the set of points in $\gX$ satisfying this equality is a \emph{zero} measure set. Thus, by replacing the log-likelihood loss with the margin loss and defining its gradients in this way, it is clear that RCAD would not perturb any data points lying on or beyond the margin. 

\begin{align}
    \label{eq:subgrad}
    \partial_{\rvx} \marginlosswxy  = \left\{ \begin{array}{ll} 
        \{-y \vw\}  & \qquad  y\cdot \innerprod{\vw}{\rvx}  < \gamma   \\
        \{-\kappa \cdot y\vw \mid \kappa \in [0, 1] \} & \qquad y\cdot \innerprod{\vw}{\rvx} = \gamma \\
        \{\mathbf{0}\} & \qquad \textrm{otherwise}
    \end{array} \right.
\end{align}

The reformulated RCAD objective in \eqref{eq:mpe-constr-obj-app} is a non-convex optimization problem with convex constraints. Thus, different solvers may yield different solutions. The solver we choose to analyze is a typical one -- taking projected gradient ascent steps to maximize the entropy term, starting from an initialization $\vw^{(0)}$ that satisfies the constraint in \eqref{eq:mpe-constr-obj-app}, and one where the iterates $\witer{t}$ continue to do so after every projection step. 

Given the projection operator $\Pi_\gS$ onto the convex set $\gS$, and learning rate $\eta > 0$, an RCAD iteration's update rule for $\witer{t}$ is defined as follows:
    \begin{align*}
        \tilde{\vw}^{(t+1)} &=  \witer{t} + \eta \cdot \nabla_{\witer{t}} \gM_{\hat \gD}(\witer{t}) \\ 
        \witer{t+1} &= \Pi_{\|\vw\|_2 \leq 1} \tilde{\vw}^{(t+1)}
    \end{align*}
\textbf{Note on projection.} For simplicity we shall treat the projection operation as just renormalizing $\tilde{\vw}^{(t+1)}$ to have unit norm, $\ie$ $\|{\vw}^{(t+1)}\|_2 = 1$, $\forall t \geq 0$. This is not necessarily restrictive. In fact, the optimal solution with the lowest test error: $\vw^* \triangleq \brck{1, 0, \dots, 0}^{\top}$, satisfies $\|\vw^*\|_2^2 = 1$, and thus lies on the boundary of $\gW$. Thus, we do not lose anything by renormalizing $\tilde{\vw}^{(t+1)}$ to get the next iterate ${\vw}^{(t+1)}$.
With this, we are ready to state an informal outline of our proof strategy.

\textbf{Informal proof outline.} Our main approach to prove the claim in Theorem~\ref{thm:main-result} can be broken down and outlined in the following way:
\begin{itemize}[topsep=8pt,leftmargin=16pt]
    \item \textbf{Lemma~\ref{lem:stage-1-gen-bound}: }First, we shall rely on typical margin based generalization bounds to show that if our training loss is low $(\leq \rho/2)$ and training data size $n$ is large enough, then we can bound the test error of the ERM solution by some small value $\rho$ with high probability. 
    \item \textbf{Lemma~\ref{lem:init-cond}:} Next we shall use the fact that the test error of the ERM solution is less than $\rho$, to prove a lower bound on the true feature weights $\wone$. This tells us that the true feature has been sufficiently learned, but ERM still depends on the noisy dimensions since $\|\wtwo\|_2 \gg 0$, and this cannot be remedied by ERM without sampling more training data. 
    \item \textbf{Lemma~\ref{lem:update-cond}:} Now, given the ERM solution as the initialization we take steps of projected gradient ascent on the reformulated RCAD objective in \eqref{eq:mpe-constr-obj-app}. For this procedure, under some conditions on the noise terms, margin $\gamma$ and step size $\alpha$ we show that the partial derivative of the population objective $\gM_\gD(\vw)$ with respect $\wone$ is along $\wone$ at every step:  $\innerprod{\pardev{\gM_\gD(\witer{t})}{\woneiter{t}}}{\woneiter{t}} > 0$. On the other hand, we show that the derivative with respect to $\wtwo$ is sufficiently negatively correlated with $\wtwo$ at every step: $\innerprod{\nabla_\wtwoiter{t}\gM_\gD(\witer{t})}{\wtwoiter{t}} < -c_1\cdot \|\wtwoiter{t}\|_2^2$. Also, we note that projecting the iterates to a unit norm ball is sufficient to satisfy the constraints in \eqref{eq:mpe-constr-obj-app} after every ascent step.
    \item \textbf{Finally,} we use the above two sufficient conditions to arrive at the following conclusion: RCAD gradient ascent iterations on $\gM_\gD(\vw)$  monotonically increase the norm along the true feature $(|\woneiter{t}| \uparrow)$ and monotonically decrease the norm along the irrelevant noisy features $(\|\wtwoiter{t}\|_2 \downarrow)$, thereby unlearning the spuriousness. Next, we rely on some uniform convergence arguments from prior works~\cite{chen2020self} to show that the previous two points are true even if we approximate the population gradient $\nabla_\vw \gM_\gD(\vw)$, with the finite sample approximation $\nabla_\vw \gM_{\hat \gD}(\vw)$, yielding the high probability finite sample guarantees in Theorem~\ref{thm:main-result}. 
\end{itemize}

\subsection{Technical Lemmas}
\label{app:subsec:technical-lemmas}

In this section we shall state some technical lemmas without proof, with references to works that contain the full proof. We shall use these in the following sections when proving our lemmas in Section~\ref{sec:analysis}. 

\begin{lemma}[Lipschitz functions of Gaussians~\cite{wainwright2019high}]
\label{lem:functions-of-gaussian}
Let $X_1, \ldots, X_n$ be a vector of \iid Gaussian variables and $f:\Real^n \mapsto \Real$ be $L$-Lipschitz with respect to the Euclidean norm. Then the random variable $f(X) - \E[f(X)]$ is sub-Gaussian with parameter at most $L$, thus:
\begin{align*}
    \Prob[|f(X) - \E[f(X)]| \geq t] \leq 2\cdot \exp{\paren{-\frac{t^2}{2L^2}}}, \;\; \forall \, t\geq 0.
\end{align*}
\end{lemma}
\vspace{0.3in}

\begin{lemma}[Hoeffding bound~\cite{wainwright2019high}]
\label{lem:hoeffding}
Let $X_1, \ldots, X_n$ be a set of $\mu_i$ centered independent sub-Gaussians,  each with parameter $\sigma_i$. Then for all $t\geq 0$, we have
\begin{align*}
    \Prob \brck{\frac{1}{n}\sum_{i=1}^{n} (X_i - \mu_i) \geq t} \leq \exp\paren{-\frac{n^2t^2}{2\sum_{i=1}^{n}\sigma_i^2}}.
\end{align*}
\end{lemma}

\vspace{0.3in}
\begin{lemma}[\citet{koltchinskii2002empirical}]
\label{lem:margin-bd}
If we assume $\forall f:\calX \mapsto \Real \in \calF$, $\sup_{\rvx \in \calX} |f(\rvx)| \leq C$, then with probability at least $1-\delta$ over sample $\hat \gD$ of size $n$, given $\marginlosswxy$ as the margin loss with some $\gamma > 0$, and $\gR_n(\gF)$ as the empirical Rademacher complexity of class $\gF$, the following uniform generalization bound holds $\forall f \in \gF$, 
\begin{align*}
    \E_{\gD} \I(y\cdot \wtx < 0) \; \; \leq\;\; \E_{\hat{\gD}} \I(\marginlosswxy > 0)  + 4\frac{\gR_n(\gF)}{\gamma} + \sqrt{\frac{\log\log(4C / \gamma)}{n}} + \sqrt{\frac{\log{1/\delta}}{2n}}.
\end{align*}
\end{lemma}

\subsubsection{Some useful facts about the \texorpdfstring{$\erfc{\cdot}$}{erfc} function}

In this section we define the complementary error function: $\erfc{x}$ and look at some useful upper and lower bounds for $\erfc{x}$ when $x \geq 0$, along with the closed form expression for its derivative. 
If $\Phi(x) \triangleq \Prob_{z\sim \gN(0, 1)} (z \leq x)$ is the cumulative distribution function of a standard normal, then the complement $\Phi_c(x) = 1 - \Phi(x)$, is related to the $\textrm{erfc}$ function in the following way: 
\begin{align}
     & \Phi_c(x) \; =\;  \frac{1}{\sqrt{2\pi}} \cdot \int_{z \geq x} \exp\paren{-z^2/2}\cdot dz \nonumber \\ 
     & \erfc{x} \; =\;  2 \cdot \Phi_c(\sqrt{2}x) \;=\; \frac{2}{\sqrt{\pi}} \cdot \int_{z \geq x} \exp\paren{-z^2}\cdot dz \label{eq:erfcdefinition}
\end{align}

The $\textrm{erfc}$ function is continuous and smooth, with its derivative having a closed form expression, as given by \eqref{eq:erfcderivative}. For our calculations we would also need convenient upper and lower bounds on the $\textrm{erfc}$ function itself, given by \eqref{eq:erfcub} and \eqref{eq:erfclb} respectively. Full proofs for these bounds can be found in~\citet{kschischang2017complementary}. 

\begin{align}
    & \frac{d}{dx}\paren{\erfc{x/\sqrt{2}}} \; = \; -{\sqrt{\frac{2}{\pi}}} \cdot \exp{\paren{-x^2/2}} \label{eq:erfcderivative} \\
    & \erfc{\frac{x}{\sqrt{2}}}  \; \leq \; \sqrt{\frac{2}{\pi}} \cdot \frac{\exp{\paren{-x^2/2}}}{x},   \quad\quad\textrm{erfc upper bound when}\; x \geq 0 \label{eq:erfcub} \\
    & \erfc{\frac{x}{\sqrt{2}}} \; \geq \;   2\sqrt{\frac{2}{\pi}} \cdot \frac{ \exp{\paren{-x^2/2}}}{(x + \sqrt{x^2 + 4})},\quad\quad\textrm{erfc lower bound when}\; x \geq 0 \label{eq:erfclb}
\end{align}

Now, we are ready to begin proving our lemmas in Section~\ref{sec:analysis}. We prove them in the same order as listed in the informal proof sketch provided in Section~\ref{app:sec:proofs-mpe-analysis}.

\subsection{Proof of Lemma~\ref{lem:init-cond}}

First, we state and prove the following lemma that bounds the generalization gap of $l_2$ norm constrained linear predictors trained by minimizing a margin loss on training data.

\begin{lemma}[generalization gap] 
\label{lem:stage-1-gen-bound}
With probability $1$$-$$\delta$ over $\hat \gD$, $\forall \vw \in \gW,$
\begin{align*}
        \errD \leq \marginerrDhat  + \tilde{\gO}\paren{\frac{\beta + \sqrt{\log (1/\delta)  \|\tilde{\Sigma}\|_{\mathrm{op}}} + \sqrt{\|\tSigma\|_*}}{\gamma\sqrt{n}}},
\end{align*}
where $\tilde{\gO}$ hides $\polylog$ factors in $\beta, \gamma, \|\tilde{\Sigma}\|$. 
\end{lemma}

\emph{Proof.}

The proof is a simple application of the margin based generalization bound in Lemma~\ref{lem:margin-bd}. For this we first need to prove the claim in Proposition~\ref{prp:hp-norm-bound} which yields a high probability bound over the function $|\innerprod{\vw}{\rvx}|:\gX \mapsto \Real$ which belongs to our class $\gW$ when $\|\vw\|_2 \leq 1$. Then, we shall bound the empirical Rademacher complexity of the function class $\gW$ using Proposition~\ref{prp:rad-bd}. These two high probability bounds are used to satisfy the conditions for Lemma~\ref{lem:margin-bd}.
Finally, we use a union bound to show that Proposition~\ref{prp:hp-norm-bound}, Proposition~\ref{prp:rad-bd} along with Lemma~\ref{lem:margin-bd} hold true with with probability $\geq 1- \delta$, to present a final generalization gap. 

In the end, we arrive at a lower bound on training sample size $n$, indicating the number of samples that are sufficient for the ERM solution obtained by minimizing the margin loss over the training data to have test error at most $\rho$.

\begin{proposition}[high probability bound over $|\wtx|$]
\label{prp:hp-norm-bound}
If $\|\vw\|_2 \leq 1$, then with probability $ \geq 1- \frac{\delta}{2}$, we can bound  $|\wtx|$ using Lemma~\ref{lem:functions-of-gaussian}, 
\begin{align*}
    |\wtx| \leq \sqrt{2\opnorm{\tSigma} \cdot \log \frac{2}{\delta}} + \sqrt{\tr\paren{\tSigma}} + \beta
\end{align*}
\end{proposition}

\textit{Proof.}

Since $\ltwonorm{w} \leq 1$, by Cauchy-schwartz we get $|\wtx| \leq \|\vw\|_2 \|\rvx\|_2 \leq \|\rvx\|_2$. Now we try to get a high probability bound over $\|\rvx\|_2$. Since we know that $\rvx$ follows a multivariate Gaussian distribution, specified by \eqref{eq:analysis-sample-app}, we can use triangle inequality to conclude that $\|\rvx\|_2 \leq \beta + \|\tSigma^{1/2}\rvz\|_2$ since $\rvx = y\beta \cdot \brck{1, 0, \ldots, 0}^\top + \tSigma^{1/2}\rvz$,  where $\rvz \sim \gN(\bf{0}, \bf{I}_{d+1})$. 

Hence, all we need to do is get a high probability bound over $\|\tSigma^{1/2}\rvz\|_2$ which is a function of $d+1$ independent Gaussian variables. Thus, we can apply the concentration bound in Lemma~\ref{lem:functions-of-gaussian}. But before that, we need to compute the Lipschitz constant for the the function $g(\rvz) = \|\tSigma^{1/2}\rvz\|_2$ in the eculidean norm.
\begin{align}
    \label{eq:lip-const}
    |\|\tSigma^{1/2}\rvz_1\|_2 - \|\tSigma^{1/2}\rvz_2\|_2| \; \leq \; \|\tSigma^{1/2}(\rvz_1 - \rvz_2)\|_2 \; \leq \; \sqrt{\opnorm{\tSigma}} 
\end{align}

Using Lipschitz constant from \eqref{eq:lip-const} in Lemma~\ref{lem:functions-of-gaussian}, we arrive at the following inequality which holds with probability at least $1-\frac{\delta}{2}$.
\begin{align}
    \label{eq:hp-bd}
    |\wtx| \leq \sqrt{2\opnorm{\tSigma} \cdot \log \frac{2}{\delta}} + \E[\|\tSigma^{1/2}\rvz\|_2] + \beta
\end{align}

We shall simplify $\E\brck{\|\tSigma^{1/2}\rvz\|_2^2}$ in the following way: 
\begin{align*}
    \E\brck{\|\tSigma^{1/2}\rvz\|_2^2} = \E \brck{\tr\paren{\rvz^\top \tSigma \rvz}} =  \tr\paren{\tSigma \;\E\brck{\rvz\rvz^\top}} = \tr\paren{\tSigma}  
\end{align*}

Now, we can use Jensen: $\E\brck{\sqrt{\|\tSigma^{1/2}\rvz\|_2^2}} \leq \sqrt{\E\brck{{\|\tSigma^{1/2}\rvz\|_2^2}}} = \sqrt{\tr\paren{\tSigma}}$.
Applying this result into \eqref{eq:hp-bd}, we finish the proof for Proposition~\ref{prp:hp-norm-bound}. 

The next step in proving Lemma~\ref{lem:stage-1-gen-bound} is to bound $\gR_n(\gW)$ which is the empirical Rademacher complexity of the class of linear predictors in $d+1$ dimensions with $l_2$ norm $\leq 1$. We state an adapted form of the complexity bound (Proposition~\ref{prp:rad-bd}) for linear predictors from~\citet{kakade2008complexity} and then apply it to our specific class of linear predictors $\gW$.

\vspace{0.2in}
\begin{proposition}[$\gR_n(\gW)$ bound for linear functions~\cite{kakade2008complexity}]
\label{prp:rad-bd}
Let $\gW$ be a convex set inducing the set of linear functions $\gF(\gW) \triangleq \{\wtx: \calX \mapsto \Real \mid w \in \calW\}$ for some input space $\calX$, bounded in norm $\|\cdot\|$ by some value $R>0$. Now, if $\exists$ a mapping $h:\gW \mapsto \Real$ that is $\kappa$-strongly  with respect to the dual norm $\|\cdot\|_*$ and some subset $\gW' \subseteq \gW$ takes bounded values of $h(\cdot)$ \ie $\{h(\vw) \leq K \mid \vw \in \gW'\}$ for some $K>0$, then the empirical Rademacher complexity of the subset $\gW'$ given by $\gR_n(\gF(\gW')) \leq R\sqrt{\frac{2K}{\kappa n}}$.   
\end{proposition}

Let $\|\cdot\|_2^2$ be the function $h:\gW \mapsto \Real$ in Proposition~\ref{prp:rad-bd}, and we know that $\|\cdot\|_2^2$ is $2$-strongly convex in $l_2$ norm. Further, take the standard $l_2$ norm as the norm over $\gX$. So, the dual norm $\|\cdot\|_*$ is also given by $l_2$ norm. Thus, $\kappa = 2$. We also know that $\gW$ is bounded in $\|\cdot\|_2$ by $1$, based on our setup definition. Thus, $R=1$. 

Now, we look at $K$. While proving Proposition~\ref{prp:hp-norm-bound} we proved a high probability bound over $\|\vx\|_2$, which we shall plug into the value of $K$. Thus, from our choice of $h$ above, and the result in Proposition~\ref{prp:hp-norm-bound} we can conclude that with probability $\geq 1 - \frac{\delta}{2}:$

\begin{align}
  \gR_n(\gW) \leq  \frac{\sqrt{2\opnorm{\tSigma} \log (2/\delta)} + \sqrt{\tr\paren{\tSigma}} + \beta}{\sqrt{n}} 
\end{align}
We are now ready to plugin the bounds on $\gR_n(\gW)$ and $|\wtx|$ into the margin based generalization bound in Lemma~\ref{lem:margin-bd}, where $C$ takes the value of the bound on $|\wtx|$. We set the sample size $n$ large enough for Lemma~\ref{lem:margin-bd} to hold with probability $\geq 1 - \frac{\delta}{2}$. Thus, by union bound the following is satisfied with probability $\geq 1 - \delta$.
\begin{align}
    &\E_{\gD} \I(y\cdot \wtx < 0)  \leq \E_{\hat{\gD}} \I(\marginlosswxy > 0) + \tilde{\gO} \paren{\frac{\beta + \sqrt{\opnorm{\tSigma} \log (1/\delta)} + \sqrt{\tr\paren{\tSigma}}}{\gamma\sqrt{n}}} \label{eq:gen-bound}
\end{align}
Corollary~\ref{cor:ncond} immediately gives us a sufficient condition on $n$ for the test error of the ERM solution to be upper bounded by $\rho$. Note, that from our constraint in \eqref{eq:mpe-constr-obj-app} we know that the training error of the ERM solution is at most $\rho/2$.

\begin{corollary}[sufficient condition on $n$ for ERM test error $\leq \rho$]
\label{cor:ncond} If  $n \gsim 
\frac{\beta^2 + \log (1/\delta) \cdot \|\tilde{\Sigma}\|_{\mathrm{op}}+\|\tSigma\|_*}{\gamma^2\rho^2}$, then the test error of the ERM  solution can be bounded by $\rho$, with probability $\geq 1 - \delta$ over $\hat \gD$. 
\end{corollary}
\emph{Proof.} 

From the constraint in \eqref{eq:mpe-constr-obj-app} we know that the training loss on $\hat \gD$ is at most $\rho /2 $ \ie  $\errDhat \leq \rho/2$ in Lemma~\ref{lem:margin-bd}. If we can upper bound the generalization error by $\frac{\rho}{2}$ then the result follows. The generalization error is bounded using \eqref{eq:gen-bound}.

This completes our proof of Lemma~\ref{lem:stage-1-gen-bound}.

\newcommand{\muone}{\mu_{+}}
\newcommand{\mutwo}{\mu_{-}}

Let $\muone \triangleq \brck{\beta, 0, \dots, 0} \in \Real^{d+1}$ and $\mutwo \triangleq \brck{-\beta, 0, \dots, 0} \in \Real^{d+1}$ and $\sigmaw$ be as defined in \eqref{eq:analysis-sample-app}. If we assume that the test error for some $\vw$ is sufficiently upper bounded: $\errD \leq  \rho$, then we can show that the norm along the true feature $\wone$ is also appropriately lower bounded using the following sequence of implications:  

\begin{align*}
    & \errD \;\; \leq \;\; \rho  \\
    & \implies \frac{1}{2} \cdot \E_{\rvz \sim \gN(\bf{0}, \bf{I}_{d+1})} \I(\innerprod{\vw}{\muone + \tSigma^{1/2}\rvz} < 0) +  \frac{1}{2} \cdot \E_{\rvz \sim \gN(\bf{0}, \bf{I}_{d+1})} \I(\innerprod{\vw}{\mutwo + \tSigma^{1/2}\rvz} > 0) \leq  \rho \\
    & \implies\;\;  \frac{1}{2} \cdot \E_{\rz \sim \gN(\bf{0}, \sigmaw^2)} \I(\beta\wone + \rz  < 0) +  \frac{1}{2} \cdot \E_{\rvz \sim \gN(\bf{0}, \sigmaw^2)} \I(-\beta\wone + \rz  > 0) \; \leq \; \rho \\
    & \implies\;\; \Prob_{\rz \sim \gN(\bf{0}, \sigmaw^2)} (z > \beta \wone) \; \leq \; \rho \\
    & \implies\;\; \Prob_{\rz \sim \gN(\bf{0}, 1)} \paren{z > \frac{\beta}{\sigmaw}\cdot \wone} \; \leq \; \rho \\
    & \implies\;\; \frac{1}{2}\cdot \erfc{\frac{\beta}{\sqrt{2}\sigmaw}\cdot \wone} \; \leq \; \rho \fourquad \textrm{since,}\;\; \Phi_c(t) = \frac{1}{2}\erfc{t/\sqrt{2}} \\ 
    & \implies\;\; \wone \; \geq \; \frac{\erfcinv{2\rho}}{\beta} \cdot \sqrt{2}\sigmaw \; \geq \; \frac{\erfcinv{2\rho}\cdot \sqrt{2\sigmaMintSigma}}{\beta}  
\end{align*}

This completes our proof of Lemma~\ref{lem:init-cond}. 

From Corollary~\ref{cor:ncond} and the arguments above, we arrive at the high probability result in Lemma~\ref{lem:init-cond}, \ie if $n \gsim
\frac{\beta^2 + \log (1/\delta) \cdot \|\tilde{\Sigma}\|_{\mathrm{op}}+\|\tSigma\|_*}{\gamma^2\rho^2}$, then with probability $\geq 1 - \delta $ over $\gD'$, $\wone \geq \frac{\erfcinv{2\rho}\cdot \sqrt{2\sigmaMintSigma}}{\beta} $. 

\begin{corollary}[ERM solution can have non-negligible dependence on $\wtwo$]
For the ERM solution $\vw$, in the worst cast $\wone = \frac{\erfcinv{2\rho}\cdot \sqrt{2\sigmaMintSigma}}{\beta} < 1$ and  given $\|\vw\|_2 = 1$, it is easy to see that $\|\wtwo\|_2 \gg 0$.
\end{corollary}
\emph{Proof.} Assume $ \frac{\erfcinv{2\rho}\cdot \sqrt{2\sigmaMintSigma}}{\beta} \geq 1$. $\implies 2\rho \leq \erfc{\beta/\sqrt{2\sigmaMintSigma}}$. This further implies $\rho \leq (1/2) \cdot \erfc{\beta/\sqrt{2\sigma_1}}$ which is not possible since the optimal test error achieved by $\vw^*$ is $(1/2) \cdot \erfc{\beta/\sqrt{2\sigma_1}}$ and $\vw \neq \vw^*$.

\subsection{Proof of Lemma~\ref{lem:update-cond}}

Before proving Lemma~\ref{lem:update-cond} we derive the closed form expression for the population objective $\gM_\gD(\vw)$ in Lemma~\ref{lem:redefine-pop-obj} which we shall use repeatedly in the sections that follow.

\begin{lemma}[$\gM_\gD(\vw)$ closed form]
\label{lem:redefine-pop-obj} If $\alpha \geq \gamma$, $\|\vw\|_2=1$  and if we define 
\begin{align*}
    \sigmaw \triangleq \sqrt{\vw^\top \tSigma \vw},\;\;\;\; \aw \triangleq \frac{\beta\wone - \gamma}{\sigmaw},
\end{align*}then we can write the closed form for $\gM_\gD(\vw)$ as:
\begin{align}
      \gM_\gD(\vw) \;\; = \;\; &\frac{1}{2} \cdot \exp{\paren{\beta \wone - \alpha \ltwonorm{\vw}^2 + \frac{\sigmaw^2}{2}}} \cdot \erfc{{\frac{\sigmaw + \aw}{\sqrt{2}}}} \nonumber \\ 
    & +\frac{1}{2} \cdot \exp{\paren{-\beta \wone + \frac{\sigmaw^2}{2}}} \cdot \erfc{{\frac{\sigmaw - \aw}{\sqrt{2}}}}.
\end{align}
\end{lemma}

\textit{Proof.}

Using the definition of the subgradient $\partial_{\rvx} \marginlosswxy$ in \eqref{eq:subgrad}, the population objective can be broken down into the following integrals, where $\gD(\rvx, \ry)$ is the measure over the space $\gX \times \gY$ defined by distribution $\gD$. 

\begin{align*}
    \gM_{\gD}(\vw)  &\;\; \triangleq  \;\;  \E_{{\gD}} \; \exp(-|\innerprod{\vw}{\rvx + \alpha \cdot \nabla_{\vx} \marginlosswxy}|)  \\
    &= \int_{y\innerprod{\vw}{\rvx} < \gamma} \exp(-|\innerprod{\vw}{\rvx - \alpha y\vw}|) \cdot \calD(\rvx, \ry) \;\; + \;\; \int_{y\innerprod{\vw}{\rvx} \geq \gamma} \exp(-|\innerprod{\vw}{\rvx}|) \cdot \calD(\rvx, \ry) \\
    &= \left. \frac{1}{2} {\int_{\innerprod{\vw}{\rvx} > -\gamma} \exp(-|\vw^\top\rvx + \alpha \ltwonorm{\vw}^2|) \cdot \calD(\rvx \mid \ry = -1)} \;\;\;\; \right\} \textrm{\textcircled{1}} \\
    & \;\; + \;\; \left. \frac{1}{2}  {\int_{\innerprod{\vw}{\rvx} < \gamma} \exp(-|\vw^\top\rvx - \alpha \ltwonorm{\vw}^2|) \cdot  \calD(\rvx \mid \ry = 1)} \;\;\;\; \right\} \textrm{\textcircled{2}} \\
    & \;\; + \;\; \left. \frac{1}{2}  {\int_{\innerprod{\vw}{\rvx} \leq -\gamma} \exp(-|\vw^\top\rvx|) \cdot  \calD(\rvx \mid \ry = -1)} \;\;\;\; \right\} \textrm{\textcircled{3}} \\
    & \;\; + \;\; \left. \frac{1}{2}  {\int_{\innerprod{\vw}{\rvx}  \geq \gamma} \exp(-|\vw^\top\rvx|) \cdot  \calD(\rvx \mid \ry = 1)} \;\;\;\; \right\} \textrm{\textcircled{4}}
\end{align*}

where the final equation uses the fact that $y \sim \textrm{Unif} \{-1, 1\}$. We are also given that the measure  $\calD(\ervx_1 \mid \ry = -1)$ for the conditional distribution over $\xone$ given $y=-1$ has distribution $\gN(-\beta, \sigma_1^2)$. Similarly, $\calD(\ervx_1 \mid \ry = +1)$ follows $\gN(\beta, \sigma_1^2)$ and measure $\calD(\rvx_2)$ follows $\gN(\bf{0}, \Sigma)$.

Thus, \begin{align*}
    \wtx \mid (\ry = -1) \sim \gN(-\beta \wone, \sigmaw^2) \\ 
    \wtx \mid (\ry = 1) \sim \gN(\beta \wone, \sigmaw^2).
\end{align*}

We shall substitute the above into \textcircled{1}, \textcircled{2}, \textcircled{3}, \textcircled{4}, to get:

\begin{align*}
    &\textrm{\textcircled{1}} = \frac{1}{2} {\int_{-\beta \wone + \sigmaw z > -\gamma} \exp(-|-\beta\wone + \sigmaw z + \alpha \ltwonorm{\vw}^2|)  \;p(z)\;\; dz} \\
    &\textrm{\textcircled{2}} = \frac{1}{2} {\int_{\beta \wone + \sigmaw z < \gamma} \exp(-|\beta\wone +  \sigmaw z - \alpha \ltwonorm{\vw}^2|)  \;p(z)\;\; dz}  \\
    &\textrm{\textcircled{3}} = \frac{1}{2} {\int_{-\beta \wone + \sigmaw z \leq -\gamma} \exp(-|-\beta\wone + \sigmaw z|)\;p(z)\;\; dz}  \\
    &\textrm{\textcircled{4}} = \frac{1}{2} {\int_{\beta \wone + \sigmaw z \geq \gamma} \exp( -|\beta\wone +\sigmaw z|) \;p(z)\;\; dz}  \\
\end{align*}

where $p(z) \triangleq \frac{1}{\sqrt{2\pi}} \exp\paren{-z^2 / 2}$ is the density of a standard Gaussian random variable. Now we shall look at the region where each of the four integrals are defined and apply the inequality $\alpha \geq \gamma$, which is stated as a condition on the step size $\alpha$ in Lemma~\ref{lem:redefine-pop-obj}. Recall, the definition of $\aw = \frac{\beta \wone - \gamma}{\sigmaw}$ and the condition $\|\vw\|_2 = 1$.

\begin{align*}
    &\textrm{Region for \textcircled{1}}: \; (\alpha \geq \gamma) \; \textrm{and} \; (-\beta\wone + \sigmaw z > -\gamma) \; \implies -\beta\wone + \sigmaw z  +\alpha \|\vw\|_2^2 > 0 \;  \\
    &\textrm{Region for \textcircled{2}}: \; (\alpha \geq \gamma) \; \textrm{and} \; (\beta\wone + \sigmaw z < \gamma) \; \implies \beta\wone + \sigmaw z  - \alpha \|\vw\|_2^2 < 0 \;  \\
    &\textrm{Region for \textcircled{3}}: \; (\alpha \geq \gamma) \; \textrm{and} \; (-\beta\wone + \sigmaw z \leq -\gamma) \; \implies -\beta\wone + \sigmaw z  \leq 0 \;  \\
    &\textrm{Region for \textcircled{4}}: \; (\alpha \geq \gamma) \; \textrm{and} \; (\beta\wone + \sigmaw z \geq \gamma) \; \implies \beta\wone + \sigmaw z  \geq 0 \;  
\end{align*}

As a result, we can further simplify the expressions into:

\begin{align*}
    &\textrm{\textcircled{1}} = \frac{1}{2} {\int_{-\beta \wone + \sigmaw z > -\gamma} \exp(\beta\wone - \sigmaw z - \alpha \ltwonorm{\vw}^2|)  \;p(z)\;\; dz} \\
    &\textrm{\textcircled{2}} = \frac{1}{2} {\int_{\beta \wone + \sigmaw z < \gamma} \exp(\beta\wone + \sigmaw  z - \alpha \ltwonorm{\vw}^2)  \;p(z)\;\; dz}  \\
    &\textrm{\textcircled{3}} = \frac{1}{2} {\int_{-\beta \wone + \sigmaw z \leq -\gamma} \exp(-\beta\wone + \sigmaw z)  \;p(z)\;\; dz}  \\
    &\textrm{\textcircled{4}} = \frac{1}{2} {\int_{\beta \wone + \sigmaw z \geq \gamma} \exp( -\beta\wone - \sigmaw z )  \;p(z)\;\; dz}  \\
\end{align*}

Now we shall note that $\textrm{\textcircled{1}} = \textrm{\textcircled{2}}$ and $\textrm{\textcircled{3}} = \textrm{\textcircled{4}}$, since $z$ and $-z$ are random variables that have the same probability distribution $\gN(0, 1)$. Using the definition of $\erfc{x}$ in \eqref{eq:erfcdefinition}, we get:
\small
\begin{align*}
      \gM_\gD(\vw) = &\frac{1}{2} \Bigg[ \exp{\paren{\beta \wone - \alpha \ltwonorm{\vw}^2 + \frac{\sigmaw^2}{2}}} \cdot \erfc{{\frac{\sigmaw + \aw}{\sqrt{2}}}}  +\\ & \qquad \exp{\paren{-\beta \wone + \frac{\sigmaw^2}{2}}} \cdot \erfc{{\frac{\sigmaw - \aw}{\sqrt{2}}}}\Bigg].
\end{align*}
\normalsize
This completes the proof of Lemma~\ref{lem:redefine-pop-obj}.
In the subsections that follow we look at the derivative of the population objective $\gM_\gD(\vw)$ with respect to weights $\vw$ at each time step $t$ of RCAD's gradient ascent iterations, specifically how well the direction of the derivative is aligned/misaligned with $\vw$.

\subsubsection{Deriving and bounding \texorpdfstring{$\innerprod{\pardev{\gM_\gD(\vw)}{\wone}}{\wone}$}{grad w1}.}

Using the reformulated expression for $\gM_\gD(\vw)$ derived in Lemma~\ref{lem:redefine-pop-obj}, and given that $\|\witer{t}\|_2^2=1$ we derive the expression for $\pardev{\gM_\gD(\witer{t})}{\woneiter{t}}$ in \eqref{eq:dmw-dwone}. Whenever convenient we drop the dependence on the superscript $(t)$. The following derivation is a simple application of the chain rule where we used the following:
\begin{align*}
    \pardev{\aw}{\wone} = \frac{\beta+(\gamma - \beta \wone)\frac{\wone\sigma_1^2}{\sigmaw^2}}{\sigmaw}, \qquad
    \pardev{\sigmaw}{\wone} =  \frac{\wone \sigma_1^2}{\sigmaw}.
\end{align*}
\newcommand{\etermwoneA}{\frac{1}{2}\exp{\paren{\beta \wone + \frac{\sigmaw^2}{2} - \alpha}}}
\newcommand{\etermwoneB}{\frac{1}{2}\exp{\paren{-\beta \wone + \frac{\sigmaw^2}{2}}}}
\newcommand{\erfctermwoneA}{\erfc{(\sigmaw + \aw)/\sqrt{2}}}
\newcommand{\erfctermwoneB}{\erfc{(\sigmaw - \aw)/\sqrt{2}}}
\newcommand{\dawdwone}{\frac{\beta+(\gamma - \beta \wone)\frac{\wone\sigma_1^2}{\sigmaw^2}}{\sigmaw}}
\newcommand{\dawdwtwo}{\frac{(\gamma - \beta \wone)}{\sigmaw^2}\frac{\Sigma\wtwo}{\sigmaw}}

\newcommand{\etermwtwo}{\frac{1}{2}\exp{\paren{-\beta \wone + \frac{\sigmaw^2}{2}}}}
\newcommand{\dsigmawdwone}{\frac{\wone \sigma_1^2}{\sigmaw}}
\newcommand{\dsigmawdtwo}{\frac{\Sigma_2 \wtwo}{\sigmaw}}

\begin{align}
\pardev{\gM_\gD(\vw)}{\wone}  \;   &= \; \etermwoneA \cdot  \Bigg(
                    (\beta-2\alpha\wone)\cdot \erfctermwoneA \nonumber \\
                    &+ \dawdwone \cdot \pardev{\paren{\erfctermwoneA}}{\aw}  
                    + \wone \sigma_1^2 \cdot \erfctermwoneA \nonumber \\
                    &+  \dsigmawdwone \cdot \pardev{\paren{\erfctermwoneA}}{\sigmaw} \Bigg)  \nonumber\\      
                    &+ \etermwoneB \cdot  \Bigg(
                    -\beta \cdot \erfctermwoneB \nonumber \\
                    &+ \dawdwone \cdot \pardev{\paren{\erfctermwoneB}}{\aw}  
                    + \wone \sigma_1^2 \cdot \erfctermwoneB \nonumber \\
                    &+  \dsigmawdwone \cdot \pardev{\paren{\erfctermwoneB}}{\sigmaw} \Bigg) 
\end{align}

Substituting the partial derivatives for $\erfc{\cdot}$ from \eqref{eq:erfcderivative} we can rewrite the above equation:

\begin{align}
\pardev{\gM_\gD(\vw)}{\wone}  \;   &= \; \etermwoneA \cdot  \Bigg(
                    (\beta-2\alpha\wone+\wone \sigma_1^2)\cdot \erfctermwoneA \nonumber \\
                    &- \sqrt{\frac{2}{\pi}} \cdot \frac{(\beta + (\gamma - \beta \wone)\frac{\wone\sigma_1^2}{\sigmaw^2} + \wone \sigma_1^2)}{\sigmaw} \cdot  \exp{\paren{-\frac{(\sigmaw + \aw)^2}{2}}} \Bigg)  \nonumber\\      
                    &+ \etermwoneB \cdot  \Bigg(
                    (-\beta + \wone\sigma_1^2) \cdot \erfctermwoneB \nonumber \\
                    & + \sqrt{\frac{2}{\pi}} \cdot \frac{(\beta + (\gamma - \beta \wone)\frac{\wone\sigma_1^2}{\sigmaw^2}- \wone \sigma_1^2)}{\sigmaw} \cdot  \exp{\paren{-\frac{(\sigmaw - \aw)^2}{2}}} \Bigg)     
            \label{eq:dmw-dwone}
\end{align}

Now we shall bound the term $\innerprod{\wone}{\pardev{\gM_\gD(\vw)}{\wone}}$ and show that $\innerprod{\wone}{\pardev{\gM_\gD(\vw)}{\wone}} > 0$ with high probability over the dataset $\gD'$. 

From our assumption on the margin term $\gamma$ in Theorem~\ref{thm:main-result} we can bound $\aw \leq c_0$, where $c_0 > 0$,
\begin{align}
 & \gamma + c_0 \sigmaMintSigma \geq \beta 
 \;\; \implies \;\;  \gamma - \beta \wone \geq -c_0 \sigmaw \implies \aw \leq c_0  \label{eq:aw-ub} \\
 \;\; \fourquad &\implies \;\;  (\beta \wone - \gamma)\frac{\wone\sigma_1^2}{\sigmaw^2} \leq c_0 \cdot {\frac{\sigma_1^2}{\sqrt{\sigmaMintSigma}}} \nonumber
\end{align}

Thus, when $\beta$ is large enough such that $\beta \gsim \sigma_1^2$, we get $(\beta + (\gamma - \beta \wone)\frac{\wone\sigma_1^2}{\sigmaw^2}- \wone \sigma_1^2) > 0$.
This is exactly the assumption regarding the separation between the class means $(\beta)$ made in Theorem~\ref{thm:main-result}, \ie we are given that $\beta \gsim \opnorm{\tSigma} \implies \beta \gsim \sigma_1^2$. Finally, we know that $\wone > 0$ from our initialization condition in Lemma~\ref{lem:init-cond}. Thus, when showing $\innerprod{\wone}{\pardev{\gM_\gD(\vw)}{\wone}} > 0$, we can ignore corresponding terms in \eqref{eq:dmw-dwone}, and it is sufficient to show:

\begin{align}
 & \exp{\paren{\beta\wone - \alpha}} \cdot  \Bigg(
                    (\beta\wone-2\alpha\wone^2+\wone^2 \sigma_1^2)\cdot \erfctermwoneA \nonumber \\
                    & \qquad \fourquad - \sqrt{\frac{2}{\pi}} \cdot \frac{(\beta\wone + (\gamma - \beta \wone)\frac{\wone^2\sigma_1^2}{\sigmaw^2}+ \wone^2 \sigma_1^2)}{\sigmaw} \cdot  \exp{\paren{-\frac{(\sigmaw + \aw)^2}{2}}} \Bigg)  \nonumber\\      
                    &+ \exp{\paren{-\beta\wone}} \cdot  \Bigg(
                    (-\beta\wone + \wone^2\sigma_1^2) \cdot \erfctermwoneB \Bigg)  > 0
                    \label{eq:pre-main-condition}
\end{align}

First, we will show that $\aw + \sigmaw \geq 0$. Then, we will prove the critical condition in \eqref{eq:main-condition}. This would imply that the term multiplied with $\exp{\paren{\beta\wone - \alpha}}$ is positive, and since for large enough $\beta$, we get $\exp{\paren{\beta\wone - \alpha}} \gg \exp{\paren{-\beta\wone}}$, we can recover the inequality in \eqref{eq:pre-main-condition}.

\begin{align}
                    &(\beta\wone-2\alpha\wone^2+\wone^2 \sigma_1^2)\cdot \erfctermwoneA \nonumber \\
                    & \qquad - \sqrt{\frac{2}{\pi}} \cdot \frac{(\beta\wone + (\gamma - \beta \wone)\frac{\wone^2\sigma_1^2}{\sigmaw^2}+ \wone^2 \sigma_1^2)}{\sigmaw} \cdot  \exp{\paren{-\frac{(\sigmaw + \aw)^2}{2}}} > 0
            \label{eq:main-condition}
\end{align}

From Theorem~\ref{thm:main-result}, we are given that $n \gsim \frac{\beta^2 + \log (1/\delta) \cdot \|\tilde{\Sigma}\|_{\mathrm{op}}+\|\tSigma\|_*}{\gamma^2\erfc{K\alpha/\sqrt{2\sigmaMintSigma}}^2} $. Applying Corollary~\ref{cor:ncond} we get that $\beta \wone \geq K \alpha$ at initialization using the following sequence of arguments:

\begin{align}
     & n \;\;\gsim \;\; \frac{\beta^2 + \log (1/\delta) \cdot  \|\tilde{\Sigma}\|_{\mathrm{op}}+\|\tSigma\|_*}{\gamma^2\erfc{K\alpha/\sqrt{2\sigmaMintSigma}}^2} \;\; \implies \;\; \rho \leq \erfc{\frac{K\alpha}{\sqrt{2\sigmaMintSigma}}} \nonumber \\
    &\qquad \implies \;\; \erfcinv{\rho} \sqrt{2\sigmaMintSigma} \geq K\alpha  
     \;\; \implies \;\; \beta \wone \geq K\alpha \label{eq:betawone-cond} 
\end{align}

Now, $\aw + \sigmaw = \frac{\beta\wone - \gamma}{\sigmaw} + \sigmaw$. Since, $\beta \wone \geq K \alpha$ for a sufficiently large $K$ that we can choose, and given that the margin $\gamma \leq \alpha$, it is easy to see that $\aw + \sigmaw > 0$. 
Thus, we can substitute $\erfctermwoneA$ with its lower bound from \eqref{eq:erfclb} into \eqref{eq:main-condition} to get:
\begin{align}
                    &(\beta\wone-2\alpha\wone^2+\wone^2 \sigma_1^2)\cdot 2\sqrt{\frac{2}{\pi}}\cdot \frac{\exp{\paren{-(\aw + \sigmaw)^2/2}}}{\aw + \sigmaw + \sqrt{(\aw + \sigmaw)^2 + 4}} \nonumber \\
                    & \qquad - \sqrt{\frac{2}{\pi}} \cdot \frac{(\beta\wone + (\gamma - \beta \wone)\frac{\wone^2\sigma_1^2}{\sigmaw^2}+ \wone^2 \sigma_1^2)}{\sigmaw} \cdot  \exp{\paren{-\frac{(\sigmaw + \aw)^2}{2}}} \label{eq:condone}
\end{align}
Simplifying the above expression by taking out common positive terms, we conclude that it is sufficient to prove the following lower bound, to effectively prove \eqref{eq:main-condition}. 
\begin{align*}
    \frac{2(\beta\wone-2\alpha\wone^2+\wone^2 \sigma_1^2)}{\aw + \sigmaw + \sqrt{(\aw + \sigmaw)^2 + 4}} \;\; -\;\;  \frac{(\beta\wone + (\gamma - \beta \wone)\frac{\wone^2\sigma_1^2}{\sigmaw^2}+ \wone^2 \sigma_1^2)}{\sigmaw} 
\end{align*}

\newcommand{\wonelsim}{\erfcinv{2\rho}\cdot \sqrt{2\sigmaMintSigma}}
Recall the margin assumption in Theorem~\ref{thm:main-result}, and the corresponding implication in \eqref{eq:aw-ub}. Thus, we know that $\aw \leq c_0$, since $\aw \leq \frac{\beta- \gamma}{\sqrt{\sigmaMintSigma}}$. Thus, $ \frac{\aw + \sigmaw + \sqrt{(\aw + \sigmaw)^2 + 4}}{2\sigmaw} \; \leq \; \tilde{c} \; \triangleq \; \frac{c_0 + \sqrt{\opnorm{\tSigma}} + \sqrt{(c_0 + \sqrt{\opnorm{\tSigma}})^2+4}}{2\sqrt{\sigmaMintSigma}}$. Let $\beta \wone \geq K\alpha \geq \wonelsim$, which is true with high probability at initialization, as we derived in \eqref{eq:betawone-cond}. Thus for some arbitrary constant $10/9$ that need only be slightly greater than $1$, if  
\begin{align}
    & \gamma \; < \; \erfcinv{2\rho}\cdot \sqrt{2\sigmaMintSigma} - \beta^2\paren{1-\frac{1}{(10/9)\tilde{c}}}\cdot \frac{\opnorm{\tSigma}}{\sigma_1^2 \paren{\wonelsim}} \label{eq:gammabenignone} \\
    & \implies \gamma \; < \; \beta\wone - \beta^2\paren{1-\frac{1}{(10/9)\tilde{c}}}\cdot \frac{\opnorm{\tSigma}}{\sigma_1^2 \paren{\wonelsim}} \nonumber \\
    & \implies \gamma \; < \; \beta\wone -  \beta\paren{1-\frac{1}{(10/9)\tilde{c}}}\cdot \frac{\opnorm{\tSigma}}{\sigma_1^2 \wone}  \; < \; \beta\wone - \beta\paren{1-\frac{1}{(10/9)\tilde{c}}}\cdot\frac{\sigmaw^2}{\wone\sigma_1^2} \nonumber \\
    & \implies \frac{\beta\wone - \gamma}{\sigmaw}\cdot \frac{\wone \sigma_1^2}{\sigmaw} \; >\; \beta \paren{1-\frac{1}{(10/9)\tilde{c}}} \nonumber \\
    & \implies \wone \cdot \frac{\beta\wone - \gamma}{\sigmaw}\cdot \frac{\wone \sigma_1^2}{\sigmaw} \; >\; \wone \cdot \beta \paren{1-\frac{1}{(10/9)\tilde{c}}} \nonumber \\
    & \implies \beta\wone - \aw \frac{\sigma_1^2 \wone^2}{\sigmaw} \; < \;\frac{9}{10}\frac{\beta\wone}{\tilde{c}} \label{eq:gammabenigntwo} 
\end{align}

Using the definition of $\tilde{c}$ and \eqref{eq:gammabenigntwo}, we can derive the following:
\begin{align}
    & \paren{\frac{2\beta\wone}{\aw + \sigmaw + \sqrt{(\aw + \sigmaw)^2 + 4}} -  \frac{(\beta\wone + (\gamma - \beta \wone)\frac{\wone^2\sigma_1^2}{\sigmaw^2}}{\sigmaw} } \nonumber \\
    &  \quad \geq \frac{2}{\aw + \sigmaw + \sqrt{(\aw + \sigmaw)^2 + 4}} \cdot\paren{\beta \wone - \tilde{c}\paren{\beta\wone - \aw\frac{\wone^2\sigma_1^2}{\sigmaw}}} \nonumber \\
    &  \quad \geq \frac{2}{\aw + \sigmaw + \sqrt{(\aw + \sigmaw)^2 + 4}} \cdot\paren{\beta \wone - \tilde{c}\frac{9}{10}\frac{\beta\wone}{\tilde{c}}} \geq c_3 \cdot \beta \wone \label{eq:condfinal}
\end{align}
for some constant $c_3 > 0$.

Now, let us come back to \eqref{eq:condone}. Under some conditions on $\gamma$ in \eqref{eq:gammabenignone}, we arrived at the final result in \eqref{eq:condfinal}. This, further implies that \eqref{eq:condone} is strictly greater than zero, when $\beta \wone \geq K \alpha$ and $\beta \gsim \opnorm{\tSigma}$, where the former is proven and the latter is assumed. This finally recovers the critical condition in \eqref{eq:main-condition}. 

Before, we look at proving the final bound in \eqref{eq:pre-main-condition}, let us quickly revisit the condition on $\gamma$ in \eqref{eq:gammabenignone} that is sufficient to make the  claims above. If we allow $n \gsim \frac{\beta^2 + \log (1/\delta) \cdot \|\tilde{\Sigma}\|_{\mathrm{op}}+\|\tSigma\|_*}{\gamma^2\rho^2}$, then we can allow $\rho$ to be as small as we want for the right hand side in \eqref{eq:gammabenignone} to be positive. In addition, we can also control the term $\opnorm{\tSigma}$ by assuming that it is sufficiently bounded. Thus, it is not hard to see that for any large enough $\beta$, if the training data is large enough and/or noise is bounded, then there exists a valid value of the margin $\gamma$ satisfying  \eqref{eq:gammabenignone}. Hence, in a way this condition on the margin is quite benign compared to the assumption in Theorem~$\ref{thm:main-result}$, which yielded the upper bound over $\aw$ as a consequence of \eqref{eq:aw-ub}.

Given that the bound in \eqref{eq:main-condition} is satisfied, we can easily verify that \eqref{eq:pre-main-condition} would also be true since $\beta \wone \geq K\alpha$ and $\beta \gsim \opnorm{\tSigma}$. Thus, $\exists K_0$ that is large enough so that for any constant $c''' > 0$ and $\alpha > 0$:

\begin{align*}
    \exp{\paren{(K-1)\alpha}}\cdot c''' -  K\alpha \cdot \exp{\paren{-K\alpha}} > 0, \;\;\; \forall K \geq K_0
\end{align*}

This completes the proof for the first part of Lemma~\ref{lem:update-cond}, \ie we have shown $\innerprod{\wone}{\pardev{\gM_\gD(\vw)}{\wone}} > 0$ at initialization, and for any $\woneiter{t} \geq \woneiter{0}$. We will see in the proof of Theorem~\ref{thm:main-result} that this inequality is also true, since $|\wone|$ increases monotonically whenever $\innerprod{\wone}{\pardev{\gM_\gD(\vw)}{\wone}} > 0$. Thus, by an induction argument
$\innerprod{\woneiter{t}}{\pardev{\gM_\gD(\witer{t})}{\woneiter{t}}} > 0, \; \forall t \geq 0$. In the following section we look at the second part of Lemma~\ref{lem:update-cond}, involving $\nabla_{\wtwo} \gM_\gD(\vw)$.

\subsubsection{Deriving and bounding \texorpdfstring{$\innerprod{\nabla_{\wtwo} \gM_\gD(\vw)}{\wtwo}$}{grad w2}.}

Using the reformulated expression for $\gM_\gD(\vw)$ derived in Lemma~\ref{lem:redefine-pop-obj}, and given that $\|\witer{t}\|_2^2=1$ we derive the expression for $\nabla_{\wtwo} \gM_\gD(\vw)$ in \eqref{eq:dmw-dwtwo}. The following derivation is a simple application of the chain rule where we used the following:
\begin{align*}
    \nabla_{\wtwo} \aw = \frac{(\gamma - \beta \wone)}{\sigmaw}\cdot \frac{\Sigma \wtwo}{\sigmaw^2}, \qquad
    \nabla_{\wtwo} \sigmaw =  \frac{\Sigma\wtwo}{\sigmaw}.
\end{align*}
\begin{align}
    \nabla_{\wtwo}\gM_\gD(\vw)  &= \etermwoneA \cdot  \Bigg( \erfctermwoneA \cdot (\Sigma - 2\alpha \bf{I}_d)\wtwo  \nonumber \\ 
      & \fourquad+  \pardev{\paren{\erfctermwoneA}}{\sigmaw} \cdot \frac{\Sigma \wtwo}{\sigmaw}    \nonumber \\
    & \fourquad+  \pardev{\paren{\erfctermwoneA}}{\aw} \cdot \dawdwtwo   \Bigg) \nonumber \\
      & \quad + \etermwoneB \cdot \Bigg( \erfctermwoneB \nonumber \\
      & \fourquad + \frac{1}{\sigmaw} \cdot  \pardev{\paren{\erfctermwoneB}}{\sigmaw} \nonumber \\
      & \fourquad - \frac{\aw}{\sigmaw^2} \cdot  \pardev{\paren{\erfctermwoneB}}{\aw}
      \Bigg) \cdot  \Sigma\wtwo
\end{align}
Substituting the partial derivatives for $\erfc{\cdot}$ from \eqref{eq:erfcderivative} we can rewrite the above equation:
\begin{align}
    \nabla_{\wtwo}\gM_\gD(\vw)  &= \etermwoneA \cdot  \Bigg( \erfctermwoneA \cdot (\Sigma - 2\alpha \mathbf{I}_d)\wtwo  \nonumber \\ 
      & \quad - \sqrt{\frac{2}{\pi}} \exp{\paren{-\frac{(\sigmaw + \aw)^2}{2}}}  \cdot \frac{\Sigma \wtwo}{\sigmaw} + \sqrt{\frac{2}{\pi}} \exp{\paren{-\frac{(\sigmaw + \aw)^2}{2}}}  \cdot \frac{\aw\Sigma \wtwo}{\sigmaw^2}   \Bigg) \nonumber \\
      & \quad + \etermwoneB \cdot \Bigg( \erfctermwoneB \nonumber \\
      & \quad - \frac{1}{\sigmaw}  \sqrt{\frac{2}{\pi}} \exp{\paren{-\frac{(\sigmaw - \aw)^2}{2}}}  
      - \frac{\aw}{\sigmaw^2}\sqrt{\frac{2}{\pi}} \exp{\paren{-\frac{(\sigmaw - \aw)^2}{2}}} \Bigg) \cdot  \Sigma\wtwo \label{eq:dmw-dwtwo}
\end{align}

\begin{lemma}[$\|\nabla_{\wtwo}\gM_\gD(\vw)\|_2$ is bounded]
\label{lem:dmw-dwtwo-norm-bound}
Given the expression for $\nabla_{\wtwo}\gM_\gD(\vw)$ in \eqref{eq:dmw-dwtwo}, when  $\|\vw\|_2 = 1$, and the conditions on $\gamma, \alpha$ in Theorem~\ref{thm:main-result} hold true then $\|\nabla_{\wtwo}\gM_\gD(\vw)\|_2^2 \leq c_1' \|\wtwo\|_2^2$ for some $c_1' > 0$.
\end{lemma}
\emph{Proof.} 

Since $\|\vw\|_2 = 1$, for any fixed value of $\beta$ and $\alpha$, it is easy to see from $\eqref{eq:dmw-dwtwo}$ that the upper bound on $\innerprod{\nabla_{\wtwo}\gM_\gD(\vw)}{\nabla_{\wtwo}\gM_\gD(\vw)}$, will only have terms involving some positive scalar times $\|\wtwo\|_2^2$, where the scalar depends on $\sigmaMintSigma, \opnorm{\tSigma}, \beta, \alpha, \gamma$ and $c_0$. Since these are finite and positive, and $\textrm{erfc}$  is bounded, the scalar is bounded above for any value taken by $\vw$. From this, we can conclude that $\|\nabla_{\wtwo}\gM_\gD(\vw)\|_2^2 \leq c_1' \|\wtwo\|_2^2$ for some $c_1' > 0$ that is independent of $\vw$.

Now, we look at $\innerprod{\wtwo}{ \nabla_{\wtwo}\gM_\gD(\vw)}$ and show that $\exists c_1 > 0$, such that $\innerprod{\wtwo}{ \nabla_{\wtwo}\gM_\gD(\vw)} < -c_1 \cdot \|\wtwo\|_2^2$.
Since, $\textrm{erfc}$ and $\textrm{exp}$ take non-negative values, and $\Sigma$ is positive semi-definite covariance matrix, we note that:
\begin{align}
    - \sqrt{\frac{2}{\pi}}  \cdot \Bigg(  & \etermwoneA  \cdot \frac{\wtwo^\top \Sigma \wtwo}{\sigmaw} \cdot \exp{\paren{-\frac{(\sigmaw + \aw)^2}{2}}} \nonumber \\
    &+  \etermwoneB  \cdot \frac{\wtwo^\top \Sigma \wtwo}{\sigmaw}  \cdot \exp{\paren{-\frac{(\sigmaw - \aw)^2}{2}}} \Bigg) < 0 \nonumber
\end{align}
Recall \eqref{eq:betawone-cond} where we derived that $\beta\wone \geq K$ for some large $K > 0$, and we assume $K\alpha \geq K \gamma$ in Theorem~\ref{thm:main-result}. Thus, $\aw = \frac{\beta\wone - \gamma}{\sigmaw} > 0$, with high probability given sufficient training samples.  Hence,
\begin{align*}
    - \sqrt{\frac{2}{\pi}} \cdot \etermwoneB  \cdot \frac{\aw}{\sigmaw^2} \cdot \wtwo^\top \Sigma \wtwo \cdot \exp{\paren{-\frac{(\sigmaw - \aw)^2}{2}}} < 0
\end{align*}
Thus, the only remaining terms are:
\begin{align}
    & \etermwoneA  \cdot \Bigg( \erfctermwoneA \cdot \wtwo^\top (\Sigma - 2\alpha \mathbf{I}_d)\wtwo \nonumber\\ 
      &  + \sqrt{\frac{2}{\pi}} \exp{\paren{-\frac{(\sigmaw + \aw)^2}{2}}} \cdot \frac{\aw}{\sigmaw^2} \wtwo^\top\Sigma\wtwo \Bigg)  \nonumber \\
      & + \etermwoneB \cdot \erfctermwoneB  \cdot  \wtwo^\top\Sigma\wtwo  
      \label{eq:wtworemaining}
\end{align}
Since $\aw + \sigmaw > 0$, we use the upper bound from \eqref{eq:erfcub} on $\erfctermwoneA$. Thus, we realize that if \eqref{eq:wtworemainingtwo} holds true, then we get the desired result:  $\innerprod{\wtwo}{ \nabla_{\wtwo}\gM_\gD(\vw)} \leq -c_1 \|\wtwo\|_2^2$.

\begin{align}
    & \etermwoneA  \cdot \sqrt{\frac{2}{\pi}}\cdot \exp{\paren{-\frac{(\sigmaw + \aw)^2}{2}}} \cdot \paren{ \wtwo^\top \paren{\Sigma \paren{1 + \frac{\aw}{\sigmaw^2}} - 2\alpha \mathbf{I}_d} \wtwo  } \nonumber\\ 
      & + \etermwoneB \cdot \erfctermwoneB  \cdot  \wtwo^\top\Sigma\wtwo  \leq -c_1 \|\wtwo\|_2^2
      \label{eq:wtworemainingtwo}
\end{align}

Recall our assumption on the step size $\alpha$ in Theorem~\ref{thm:main-result}: $\alpha \gsim \opnorm{\tSigma}$. 
Also, from our assumption on margin $\gamma$ in Theorem~\ref{thm:main-result}, we know that $\gamma + c_0 \sigmaMintSigma \geq \beta$, which translates to $\aw \leq c_0$ using \eqref{eq:aw-ub}. Since $\beta \gsim \opnorm{\tSigma}$, we get $\gamma + c_0 \sigmaMintSigma \gsim \opnorm{\tSigma}$.  When, $\sigmaMintSigma$ is lower bounded by some positive constant $c_4$, then $\frac{1}{2} \paren{\opnorm{\Sigma} + \frac{c_0}{\sigmaMintSigma}} \lsim  \paren{c_0\sigmaMintSigma + \frac{c_0\sigmaMintSigma}{c_4^2}} \lsim \opnorm{\tSigma}$. The margin condition on $\gamma$ trivially implies this lower bound. Still, let us discuss why this is not a restricting, rather a necessary condition.

In order for RCAD to identify and unlearn noisy dimensions, it is needed that the feature variance is not diminishing. When $\sigmaMintSigma \rightarrow 0$, then both the sampled data, and the self-generated examples by RCAD would have point mass distributions, along certain directions, which can result in some degenerate cases. Essentially we are assuming the covariance matrix $\Sigma$ to be positive definite. This is not a strong assumption and is needed to provably unlearn components along the $d$ noisy dimensions. If the variance is $0$ along some direction, and if the initialization for ERM/RCAD has a non-zero component along that direction, there is no way to reduce this component to lower values with zero mean zero variance features along this direction. Thus, $\Sigma \succ \bf{0}$. In the case where $\sigma_1=0$, the term $\sigmaw^2$ would be vanishingly close $0$  only when $\|\vw_2\| \rightarrow 0$ \ie we are arbitrarily close to $\vw^*$.

By choosing $\alpha \gsim \opnorm{\tSigma}$, we can ensure that $\paren{\Sigma \paren{1 + \frac{\aw}{\sigmaw^2}} - 2\alpha \mathbf{I}_d}$ is a negative definite matrix with all negative singular values. Thus, we can show that $\exists c'', c''' > 0$ wherein, the first term in \eqref{eq:wtworemainingtwo} scales as $-\exp{\paren{(K-1)\alpha}}\cdot c'' \|\wtwo\|_2^2$ and the second term scales as $c''' \|\wtwo\|_2^2 \cdot \exp{\paren{-K\alpha}}$.
Now, recall \eqref{eq:betawone-cond} where we derived $\beta \wone \geq K\alpha$ and $\beta \gsim \opnorm{\tSigma}$. Thus, $\exists K_0$ that is large enough so that for any constants $c_1, c'', c''' > 0$ and $\alpha > 0$:
\begin{align*}
    -\exp{\paren{(K-1)\alpha}}\cdot c'' \|\wtwo\|_2^2 +  c''' \|\wtwo\|_2^2 \cdot \exp{\paren{-K\alpha}} \leq -c_1 \|\wtwo\|_2^2,  \;\;\; \forall K \geq K_0 
\end{align*}

Thus, we conclude that $\innerprod{\wtwo}{ \nabla_{\wtwo}\gM_\gD(\vw)} \leq -c_1 \cdot \|\wtwo\|_2^2$, for some $c_1 >0$.
This, completes our proof for the second part of Lemma~\ref{lem:update-cond}, and by making the same induction argument that we saw at the end of the previous section, we get $\innerprod{\wtwoiter{t}}{ \nabla_{\wtwoiter{t}}\gM_\gD(\witer{t})} \leq -c_1 \cdot \|\wtwoiter{t}\|_2^2, \; \forall t\geq 0$. This completes our proof of Lemma~\ref{lem:update-cond}.

\subsection{Proof of Theorem~\ref{thm:main-result}}

From the previous sections we concluded the following results with respect to the population objective $\gM_\gD(\vw)$: \textit{(i)} $\innerprod{\pardev{\gM_\gD(\witer{t})}{\woneiter{t}}}{\woneiter{t}} > 0$; and  \textit{(ii)} $\innerprod{\nabla_\wtwoiter{t}\gM_\gD(\witer{t})}{\wtwoiter{t}} < -c_1\cdot \|\wtwoiter{t}\|_2^2$. Now, Lemma~\ref{lem:population-main-theorem} below tells us that there is an appropriate learning rate $\eta$ such that if we do projected gradient ascent on the population objective $\gM_\gD(\vw)$, starting from the initialization satisfying Lemma~\ref{lem:init-cond}, then the iterate $\witer{T}$ would be $\epsilon$ close to $\vw^*$ in $T = \gO\paren{\log\paren{\frac{1}{\epsilon}}}$ iterations.

\begin{lemma}
\label{lem:population-main-theorem} If Lemma~\ref{lem:update-cond} is true, then $\exists \eta$, such that there is a constant factor decrease in $\|\wtwoiter{t}\|_2^2$ at each time step $t$, if we perform gradient ascent on the objective $\gM_\gD(\vw)$ starting from the initialization in Lemma~\ref{lem:init-cond}. Thus, in $T = \gO\paren{\log\paren{\frac{1}{\epsilon}}}$ iterations we get:
\begin{align*}
|\woneiter{T}|\geq\sqrt{1-\epsilon^2},\; \textrm{and}\;\; \|\wtwoiter{T}\|_2 \leq \epsilon.
\end{align*}
\end{lemma}

\emph{Proof:} 

Since, $\tilde{\vw}^{(t+1)} =  \witer{t} + \eta \cdot \nabla_{\witer{t}} \gM_{ \gD}(\witer{t})$:
\begin{align*}
    \|\tilde{\vw}_2^{(t+1)}\|_2^2 \;&=\; \|\vw_2^{(t)} + \eta \nabla_{\wtwoiter{t}} \gM_\gD(\witer{t}) \|_2^2  \\
    &=\; \|\vw_2^{(t)}\|_2^2 + \eta^2 \|\nabla_{\wtwoiter{t}} \gM_\gD(\witer{t})\|_2^2 + 2\eta \innerprod{\nabla_{\wtwoiter{t}} \gM_\gD(\witer{t})}{\wtwoiter{t}} \\
    &\leq \; (1-2\eta c_1) \cdot \|\vw_2^{(t)}\|_2^2 + \eta^2 \|\nabla_{\wtwoiter{t}} \gM_\gD(\witer{t})\|_2^2, \qquad c_1>0 \textrm{ from Lemma~\ref{lem:update-cond}}
\end{align*}
In Lemma~\ref{lem:dmw-dwtwo-norm-bound} we had identified that $\|\nabla_{\wtwoiter{t}} \gM_\gD(\witer{t})\|_2^2  \leq {c_1'}^2 \|\wtwoiter{t}\|_2^2$ for some $c_1'>0$ using the expression in \eqref{eq:dmw-dwtwo}. Thus,
\begin{align}
     \|\tilde{\vw}_2^{(t+1)}\|_2^2 \; \leq \; (1- 2\eta c_1 +\eta^2 {c_1'}^2) \cdot \|\wtwoiter{t}\|_2^2 
\end{align}
Thus, for a specific choice of $\eta$, one where $(1- 2\eta c_1 +\eta^2 {c_1'}^2) < 1$, we get:
\begin{align}
\|\tilde{\vw}_2^{(t+1)}\|_2^2 \leq \kappa \cdot \|\wtwoiter{t}\|_2^2, \qquad \kappa < 1.
    \label{eq:main-w2}
\end{align}
Similarly, from $\wone$ update: $ \|\tilde{\evw}_1^{(t+1)}\|_2^2 = \|\wone^{(t)}\|_2^2 + \eta^2 \|\pardev{\gM_\gD(\witer{t})}{\woneiter{t}}\|_2^2 + 2\eta \innerprod{\pardev{\gM_\gD(\witer{t})}{\woneiter{t}}}{\woneiter{t}}$. 
Since $\innerprod{\pardev{\gM_\gD(\witer{t})}{\woneiter{t}}}{\woneiter{t}} > 0$, we get:
\begin{align}
    |\tilde{\evw}_1^{(t+1)}| \; >\; |{\evw}_1^{(t)}| 
    \label{eq:main-w1}
\end{align}
Recall that we re-normalize $\tilde{\vw}^{(t+1)}$,  \ie $\vw^{(t+1)} = \tilde{\vw}^{(t+1)} / \|\tilde{\vw}^{(t+1)}\|_2$. Hence, from \eqref{eq:main-w2}
and \eqref{eq:main-w1} we can conclude: $\|\wtwoiter{t+1}\|_2^2 = (1- \Omega(1)) \cdot \|\wtwoiter{t}\|_2^2$, a constant factor reduction. With a telescoping argument on $\|\wtwoiter{t+1}\|_2^2$, we would need $T=\gO(\log(1/\epsilon))$ iterations for $\|\wtwoiter{T}\|_2^2 \leq \epsilon^2$. This completes the proof. 

Lemma~\ref{lem:population-main-theorem} furnishes the guarantees in Theorem~\ref{thm:main-result} when we perform project gradient ascent with respect to the population objective $\gM_\gD(\vw)$. We discussed this setting first since it captures the main intuition and simplifies some of the calculation. In the next subsection we shall look at the finite sample case, where RCAD optimizes $\gM_{\hat \gD}(\vw)$. 

\subsubsection{Revisiting Lemma~\ref{lem:population-main-theorem} in a finite sample setting}

Note that Lemma~\ref{lem:population-main-theorem} furnished guarantees when we took projected gradient ascent steps on the population objective $\gM_\gD(\vw)$. In fact, it gave us $\lim_{T \rightarrow \infty} \|\wtwoiter{T}\|_2 = 0$ and $\lim_{T \rightarrow \infty} \woneiter{T} = 1$. But, in practice, we only have access to gradients of the finite sample objective $\gM_{\hat \gD}(\vw)$, where the expectation is taken with respect to the empirical measure over training data $\hat{\gD}$. 

We still follow the same arguments that we used to prove Lemma~\ref{lem:population-main-theorem}, but now with sample gradients that closely approximate the true gradients through some uniform convergence arguments 
over the random variable $\nabla_{\vw} \gM_{\hat \gD} (\vw)$. Then the monotonic results in Lemma~\ref{lem:update-cond} hold with high probability if we are given sufficient training data, \ie $n$ is large enough that  $\nabla_{\vw} \gM_{\hat \gD} (\vw) \approx \nabla_{\vw} \gM_{\gD} (\vw), \; \forall \vw$, along directions $\wone$ and $\wtwo$. To formalize this notion, we rely on an adapted form of Theorem C.1 in ~\citet{chen2020self}. Although the conditions in our adaptation (Theorem~\ref{thm:finite-sample-approx}) are slightly different, the proof technique is identical to that of Theorem C.1 in~\cite{chen2020self}. Hence, we only present the key ideas needed to ensure that finite sample approximation does not change the behavior in of $\witer{t}$ we saw in Lemma~\ref{lem:update-cond}, and refer interested readers to~\cite{chen2020self} for more details.

\begin{theorem}[finite sample approximation from~\citet{chen2020self}]
\label{thm:finite-sample-approx}
Assume that $\innerprod{\pardev{\gM_\gD(\vw)}{\wone}}{\wone} > 0$, $\innerprod{\nabla_{\wtwo} \gM_\gD(\vw)}{\wtwo} < -c_1\cdot \|\wtwo\|_2^2$, and $\|\nabla_{\wtwo} \gM_\gD(\vw)\|_2^2  \leq {c_1'}^2 \|\wtwo\|_2^2$, for constants $c_1, c_1' > 0$. If $n = \tilde{\gO}(\frac{d\opnorm{\tSigma}}{\epsilon^2}\log(1/\delta))$ where $\epsilon = \gO(c_1 \tau^2)$ for some small $\tau < 0.5$, then with probability $1-\delta$ over $\hat{\gD}$, the empirical and true gradients along $\wone$ and $\wtwo$ are close:
\begin{align}
    |\innerprod{\nabla_{\wone} {\gM_\gD(\vw)} - \nabla_{\wone} {\gM_{\hat \gD}(\vw)}}{\wone}| \leq \epsilon  \nonumber \\
    |\innerprod{\nabla_{\wtwo} {\gM_\gD(\vw)} - \nabla_{\wtwo} {\gM_{\hat \gD}(\vw)}}{\wtwo}| \leq \epsilon \label{eq:fsample-approx}
\end{align}
Furthermore, if the true feature at initialization is dominant \ie $|\woneiter{0}| \geq 0.5$ and if $\|\wtwoiter{t}\|_2 \geq \tau / 2$, then for $\eta = \gO(c_1/c_1'^2)$ the norm along $\wtwo$ shrinks by a constant factor: $\|\wtwoiter{t+1}\|_2^2 \leq (1-\gO(c_1^2/c_1'^2)) \|\wtwoiter{t}\|_2^2$. This continues until $\|\wtwoiter{t}\|_2 < \tau/2$ after which it stabilizes and $\|\wtwoiter{t}\|_2 \leq \tau$ is guaranteed in subsequent iterations.  
\end{theorem}

In Theorem~\ref{thm:finite-sample-approx} we allow for some error $\epsilon$ in approximating the true gradient with the finite sample approximation along directions $\wone$ and $\wtwo$. Note, that this error is much smaller $O(c_1 \tau^2)$, than the final error tolerated on the norm of $\wtwo$ which is $\tau$, when $\tau < 0.5$. Essentially, using Lemma C.1 in~\cite{chen2020self}, it is easy to show that \eqref{eq:fsample-approx} holds with probability $1-\delta$ if $n = \tilde{\gO}(\frac{d\opnorm{\tSigma}}{\epsilon^2}\log(1/\delta))$, where $\tilde{\gO}$ hides logarithmic factors in $1/\epsilon^2$. This, allows us to make the following replacements for our expressions in Lemma~\ref{lem:population-main-theorem} which considered the population version:
\begin{align*}
    &\innerprod{\pardev{\gM_\gD(\vw)}{\wone}}{\wone} > 0 \;\; \longleftrightarrow \;\; \innerprod{\pardev{\gM_{\hat \gD}(\vw)}{\wone}}{\wone} > -\epsilon \\
   &\innerprod{\nabla_{\wtwo} \gM_\gD(\vw)}{\wtwo} < -c_1\cdot \|\wtwo\|_2^2  \;\; \longleftrightarrow \;\; \innerprod{\nabla_{\wtwo} \gM_{\hat \gD}(\vw)}{\wtwo} < -c_1\cdot \|\wtwo\|_2^2 + \epsilon \\
   &\|\nabla_{\wtwo} \gM_\gD(\vw)\|_2^2  \leq {c_1'}^2 \|\wtwo\|_2^2 \;\; \longleftrightarrow \;\; \|\nabla_{\wtwo} \gM_{\hat \gD}(\vw)\|_2^2  \leq 2{c_1'}^2 \|\wtwo\|_2^2 + 2 \epsilon^2
\end{align*}
Thus, when $\eta = \gO\paren{\frac{c_1}{c_1'^2}}$, there is a constant factor decrease in $\|\tilde{\vw}\|_2^2$:
\begin{align*}
    \|\tilde{\vw}_2^{(t+1)}\|_2^2 \leq (1-\gO(c_1^2/c_1'^2)) \|\wtwo^{(t)}\|_2^2 + 2\eta^2\epsilon^2 + 2 \eta \epsilon
\end{align*} 
On the other hand since $\eta \epsilon$ is small, there is a significant increase in $|\tilde{\evw}_1|$:
\begin{align*}
    \|\tilde{\evw}_1^{(t+1)}\|_2^2 \geq  \|\evw^{(t)}\|_2^2 - 2 \eta \epsilon
\end{align*} 
Thus, any decrease in $|\evw_1|$ is relatively much smaller compared to decrease in $\|\wtwo\|_2$. After re-normalization we get:
\begin{align*}
    \|{\wtwo}^{(t+1)}\|_2^2 \leq (1-\gO(c_1^2/c_1'^2)) \|\wtwo^{(t)}\|_2^2 
\end{align*} 
Given the constant factor reduction, the number of iterations needed to drive  $\|{\wtwo}^{(t+1)}\|_2$ to be less than $\tau$ is $\gO(\log (1/\tau))$. Furthermore, in the finite sample case, we need $n=\tilde{\gO}(\frac{1}{\tau^4}\log(1/\delta))$ samples for the $\epsilon-$approximation in \eqref{eq:fsample-approx} to hold where $\epsilon = \gO(c_1\tau^2)$. We can now combine this requirement on the sample size with the lower bound on $n$ from Corollary~\ref{cor:ncond} needed to maintain the initialization conditions in Lemma~\ref{lem:init-cond} with $\beta\wone \geq K\alpha$, to get $n \gsim \frac{\beta^2 + \log (1/\delta) \cdot \|\tilde{\Sigma}\|_{\mathrm{op}}+\|\tSigma\|_*}{\gamma^2\erfc{K\alpha/\sqrt{2\sigmaMintSigma}}^2} + \frac{\log{1/\delta}}{\tau^4}$. This completes the proof for Theorem~\ref{thm:main-result}.

\vspace{0.2in}
\section{Additional Details for Experiments in Section~\ref{sec:experiment}}
\label{app:sec:additional-exp-results}

In this section, we first present technical details useful for reproducing our empirical results in Section~\ref{sec:experiment}. We also take a note of the computational resources needed to run the experiments in this paper. Then we present some tables containing absolute test performance values for some of the plots in Section~\ref{sec:experiment} where we plotted the relative performance of each method with respect to empirical risk minimization (ERM). Finally, we present some preliminary results on Imagenet classification benchmark.

\textbf{{Hyperparameter details.}}
The hyperparameters used to train ADA~\cite{volpi2018generalizing}, ME-ADA~\cite{zhao2020maximum}, and adversarial training (FGSM)~\cite{goodfellow2014explaining} are identical to those reported in the original works we cite. For label smoothing, we tune the smoothing parameter on the validation set and find $\epsilon=0.6$ to be optimal for the smaller dataset of CIFAR-100-2k, and $\epsilon=0.2$ yields the best results on other datasets. For RCAD we have two hyperparameters $\alpha, \lambda$ and find that $\alpha=0.5, \lambda=0.02$ gives good performance over all benchmarks except CIFAR-100-2k for which we use $\alpha=1.0, \lambda=0.1$. 

Unless specified otherwise, we train all methods using the ResNet-18~\cite{he2016deep} backbone, and to accelerate training loss convergence we clip gradients in the $l_2$ norm (at $1.0$)~\cite{zhang2019gradient,hardt2016train}. 
We train all models for $200$ epochs and use SGD with an initial learning rate of $0.1$ and Nesterov momentum of $0.9$, and decay the learning rate by a factor of $0.1$ at epochs $100, 150$ and $180$~\cite{devries2017improved}. We select the model checkpoint corresponding to the epoch with the best accuracy on validation samples as the final model representing a given training method. For all datasets (except CIFAR-100-2k and CIFAR-100-10k for which we used $32$ and $64$ respectively) the methods were trained with a batch size of $128$. For our experiments involving the larger network of Wide ResNet 28-10, we use the optimization hyperparameters mentioned in~\citet{zagoruyko2016wide}.

\textbf{{Computational resources.}} To run any experiment we used at most two NVIDIA GEFORCE GTX 1080Ti GPU cards. As we mention in the main paper, any run of our method RCAD takes at most $30\%$ more per-batch processing time than ERM with data augmentation on a given dataset.
All runs of RCAD with ResNet-18 backbone on CIFAR-100-2k, CIFAR-100-10k, CIFAR-100, CIFAR-10 and SVHN take less than $20$ hours, while Tiny Imagenet takes about $36$ hours.

\textbf{{Absolute test performance values.}} In Table~\ref{tab:absolute-values-adv-comparison}, we plot the test performance values for our method RCAD which maximizes entropy on self-generated perturbations, and adversarial baselines which minimize cross-entropy on adversarial examples. The plot in Figure~\ref{fig:compare-left} is derived from the results in Table~\ref{tab:absolute-values-adv-comparison}. Table~\ref{tab:absolute-values-adv-comparison-with-LS} compares the performance of RCAD with the most competitive adversarial baseline ME-ADA, when both methods are used in combination with the label smoothing regularizer. These results are also presented in Figure~\ref{fig:compare-left} and discussed in Section~\ref{subsec:mpe-comparison-with-adv}. For the plot in Figure~\ref{fig:addnl-results}a comparing baselines ERM and ME-ADA to our method RCAD with the larger network Wide ResNet 28-10~\cite{zagoruyko2016wide}, we show the test accuracy values on CIFAR-100(-2k/10k) in Table~\ref{tab:absolute-values-WRN}. Finally, Table~\ref{tab:absolute-values-cif-subset} presents the test accuracy of each method on the CIFAR-100 test set, when trained on training datasets of varying sizes. Figure~\ref{fig:compare-right} plots the same results relative to ERM. The trends observed here and in  Figure~\ref{fig:compare-right} are discussed in Section~\ref{subsec:MPE-low-data}. In all the above, we show the mean accuracy and $95\%$ confidence interval evaluated over $10$ independent runs.

\begin{table}[!ht]
    \centering
    \footnotesize
\begin{tabular}{rcccc|c}
Dataset & ERM & FGSM & ADA & ME-ADA & RCAD \\ \hline
CIFAR-100-2k &  $28.5 \pm 0.12 $ & $27.1 \pm 0.08 $ & $28.7 \pm 0.12 $ & $29.4 \pm 0.11 $ & $\bf{30.4 \pm 0.09}$ \\
CIFAR-100-10k &   $60.4 \pm 0.09 $ & $57.6 \pm 0.09 $ & $60.0 \pm 0.10 $ & $60.4 \pm 0.08 $ & $\bf{61.9 \pm 0.06}$ \\
Tiny Imagenet & $66.4 \pm 0.07 $ & $62.4 \pm 0.07 $ & $66.5 \pm 0.08 $ & $66.9 \pm 0.10 $ & $\bf{67.3 \pm 0.07}$ \\
CIFAR-10 &  $95.2 \pm 0.04 $ & $93.1 \pm 0.04 $ & $95.2 \pm 0.03 $ & $\bf{95.6 \pm 0.05}$ & $\bf{95.7 \pm 0.04}$ \\
CIFAR-100 &  $76.4 \pm 0.05 $ & $73.2 \pm 0.05 $ & $76.5 \pm 0.05 $ & $77.1 \pm 0.07 $ & $\bf{77.3 \pm 0.05}$ \\
SVHN &  $97.5 \pm 0.02 $ & $95.4 \pm 0.02 $ & $97.5 \pm 0.02 $ & $97.4 \pm 0.04 $ & $\bf{97.6 \pm 0.02}$ 
\end{tabular} 
\vspace{0.1in}
\caption{\footnotesize 
\textbf{RCAD consistently outperforms adversarial baselines.} In Section~\ref{subsec:mpe-comparison-with-adv} we investigated the performance of RCAD against baseline methods that use the adversarially generated perturbations differently, \ie methods that reduce cross-entropy loss on adversarial examples. This table presents the test accuracies for RCAD and other baselines with $95\%$ confidence interval computed over 10 independent runs. Figure~\ref{fig:compare-left} plots the results reported here by looking at the performance improvement of each method over the ERM baseline.}
\label{tab:absolute-values-adv-comparison}
\end{table}

\begin{table}[!ht]
\centering
\footnotesize
\setlength{\tabcolsep}{2pt}
\begin{subtable}[b]{0.44\textwidth}
\begin{tabular}{rcc}
Dataset & ME-ADA+LS & RCAD+LS \\ \hline
CIFAR-100-2k &    $30.5 \pm 0.13 $ &  $\bf{32.2 \pm 0.11  }$ \\
CIFAR-100-10k &     $61.15 \pm 0.07 $  &  $\bf{63.1 \pm 0.08  }$ \\
Tiny Imagenet &   $67.1 \pm 0.08 $ &  $\bf{67.9 \pm 0.07  }$ \\
CIFAR-10 &    $\bf{95.65 \pm 0.04 }$  &  $\bf{95.7 \pm 0.03 } $ \\
CIFAR-100 &    $77.8 \pm 0.06 $ &  $\bf{79.1 \pm 0.06  }$ \\
SVHN &    $97.42 \pm 0.04 $  &  $\bf{97.6 \pm 0.02  }$ 
\end{tabular} 
\caption{\label{tab:absolute-values-adv-comparison-with-LS}}
\end{subtable}
\begin{subtable}[b]{0.48\textwidth}
\begin{tabular}{rcc|c}
Dataset & ERM & ME-ADA & RCAD \\ \hline
C100-2k & $30.2 \pm 0.10 $ & $31.8 \pm 0.07 $ & $\bf{32.3 \pm 0.08 }$ \\
C100-10k & $62.5 \pm 0.07 $ & $62.6 \pm 0.08 $ & $\bf{62.9 \pm 0.05 }$ \\
C100 & $81.2 \pm 0.07 $ & $81.0 \pm 0.07 $ & $\bf{81.4 \pm 0.06 }$ \\
\end{tabular}
\caption{\label{tab:absolute-values-WRN}}
\end{subtable}
\vspace{0.05in}
\caption{\footnotesize 
\figleft\, \textbf{RCAD outperforms adversarial baselines even when combined with existing regularizers.}  We compare the performance of RCAD and ME-ADA when both methods are trained with label smoothing (smoothing parameter $\epsilon = 0.2$). We find that label smoothing improves the performance of both methods, but the benefit of RCAD still persists. Figure~\ref{fig:compare-left} also plots the same results, relative to the performance of ERM. 
\figright\, \textbf{RCAD can outperform baselines when used with larger backbones.} Here, we train our method RCAD and baselines ERM, ME-ADA with the larger architecture Wide ResNet 28-10 on CIFAR-100 (C100), CIFAR-100-2k (C100-2k) and CIFAR-100-10k (C100-10k) benchmarks. We find that while the performance gains of RCAD over ME-ADA and ERM are higher with the ResNet-18 backbone, RCAD continues to outperform them here as well. Figure~\ref{fig:addnl-results}a also plots the same results, relative to the performance of ERM. 
}
\label{tab:additional}
\end{table}

\begin{table}[!t]
    \centering
    \footnotesize
    \begin{tabular}{rcccc|c}
Dataset size & ERM & LS & FGSM & ME-ADA & RCAD \\
\hline
500 & $11.9\pm 0.06 $ & $13.5\pm 0.08 $ & $13.2\pm 0.06 $ & $13.1\pm 0.06 $ & $\bf{14.8\pm 0.08}$ \\
1,000 & $18.9\pm 0.07 $ & $19.7\pm 0.07 $ & $18.4\pm 0.08 $ & $18.6\pm 0.08 $ & $\bf{21.3\pm 0.05}$ \\
2,000 & $28.5\pm 0.05 $ & $29.2\pm 0.07 $ & $27.1\pm 0.07 $ & $29.4\pm 0.05 $ & $\bf{30.4\pm 0.05}$ \\
5,000 & $52.2\pm 0.07 $ & $52.7\pm 0.06 $ & $50.9\pm 0.07 $ & $51.4\pm 0.06 $ & $\bf{53.6\pm 0.05}$ \\
10,000 & $60.4\pm 0.07 $ & $61.5\pm 0.06 $ & $57.6\pm 0.06 $ & $60.4\pm 0.07 $ & $\bf{61.9\pm 0.06}$ \\
50,000 & $76.4\pm 0.07 $ & $77.4\pm 0.08 $ & $73.2\pm 0.08 $ & $77.1\pm 0.07 $ & $\bf{77.7\pm 0.06}$ 
\end{tabular}
\vspace{0.1in}
\caption{\footnotesize \textbf{RCAD is more effective than other existing regularizers in low data regime.} We evaluate the effectiveness of RCAD over other baseline regularizers as we decrease the size of the training data for CIFAR-100. We find that the performance gain from using RCAD is higher as training data size reduces, possibly since RCAD is more robust to spurious correlations in the training data (compared to other baselines), and deep models are more vulnerable to such correlations in the low data regime~\cite{zhang2021understanding}. Figure~\ref{fig:compare-right} also plots the same results, relative to the performance of ERM.}
\label{tab:absolute-values-cif-subset}
\end{table}

\textbf{Preliminary results on Imagenet classification.} We take a ResNet-50~\cite{he2016deep} model pretrained on Imagenet~\cite{deng2009imagenet} and measure its \emph{top-1} and \emph{top-5} test accuracy. Next, we tune this pretrained ResNet-50 model, using the optimization hyperparameters from~\citet{yun2019cutmix}. We refer to the model produced by this method as ``Finetuned with ERM''. We compare the test accuracies for these two models on the Imagenet test set with the test accuracy of the model that is obtained by finetuning the pretrained ResNet-50 model with the RCAD objective in \eqref{eq:mpe-main-eq} (``Finetuned with RCAD''). Results for this experiment can be found in Table~\ref{tab:imagenet-results}. We do not finetune RCAD hyperparameters $\alpha, \lambda$ on Imagenet, and retain the ones used  for the TinyImagenet benchmark. We faced issues running experiments for the competitive but expensive adversarial baseline ME-ADA since the GPU memory needed to run ME-ADA on Imagenet exceeded the limit on our GPU cards. Mainly for these two reasons, we note that our results on the Imagenet benchmark are preliminary. Nevertheless, the results are promising since we see an improvement in test accuracy when we finetune the pretrained model with our RCAD objective as opposed to the standard cross-entropy loss as part of the ERM objective. The improvement is more significant for top-1 test accuracy.

\begin{table}[!ht]
    \centering
    \footnotesize
    \begin{tabular}{rcc|c}
        Test Accuracy & Pretrained ResNet-50 & Finetuned with ERM & Finetuned with RCAD \\ \hline
        top-1 & $69.81 \pm 0.07 \%$ & $77.45 \pm 0.05 \%$ & $77.76 \pm 0.06 \%$ \\
        top-5 & $89.24 \pm 0.06 \%$ & $93.21 \pm 0.04 \%$ & $93.29 \pm 0.05 \%$ 
    \end{tabular}
    \vspace{0.1in}
    \caption{\textbf{Preliminary results on Imagenet classification.} We compare the performance of a pretrained ResNet-50 model, with the models obtained by finetuning the pretrained model with either the cross-entropy loss (finetuned with ERM) or the RCAD objective (finetuned with RCAD). We show the mean accuracy and $95\%$ confidence interval evaluated over $7$ independent runs.}
    \label{tab:imagenet-results}
\end{table}

\section{RCAD can Improve Robustness to Adversarial/Natural Distribution Shifts}
\label{app:sec:mpe-robustness-to-shifts}

Typically, methods for handling distribution shift are different from methods for improving test accuracy. 
Prior work has found that increasing robustness to distribution shifts tends to be at odds with increasing test accuracy: methods that are more robust often achieve lower test accuracy, and methods that achieve higher test accuracy tend to be less robust~\citep{raghunathan2019adversarial,zhang2019theoretically,Tsipras2019RobustnessMB}.
While the main aim of our experiments is to show that RCAD improves test accuracy, our next set of experiments investigate whether RCAD is any more robust to distribution shift than baseline methods.

For these experiments on distribution shift, we use the exact same hyperparameters as in the previous experiments. Better results are likely achievable by tuning the method for performance on these robustness benchmarks. By reporting results using the exact same hyperparameters, we demonstrate that the same method might both achieve high in-distribution performance and out-of-distribution performance.

\begin{table*}[!ht]
\setlength{\tabcolsep}{4pt}
\footnotesize
    \centering
    \begin{tabular}{rcccccc}
         \textbf{Method} & \multicolumn{2}{c}{\textbf{CIFAR-100-2k}} & \multicolumn{2}{c}{\textbf{CIFAR-100}} & \multicolumn{2}{c}{\textbf{CIFAR-10}} \\ 
         & {Clean} & {FGSM Attack} & {Clean} & {FGSM Attack} & {Clean} & {FGSM Attack} \\ \hline
        ERM & 28.5 $\pm$ 0.12 & 24.7 $\pm$ 0.10 & 76.4 $\pm$ 0.05 & 67.4 $\pm$ 0.07 & 95.2 $\pm$ 0.04 & 88.3 $\pm$ 0.03 \\
        Adv. Training & 27.1 $\pm$ 0.08  & 27.2 $\pm$ 0.08 & 73.2 $\pm$ 0.05 & 73.1 $\pm$ 0.04 & 93.1 $\pm$ 0.04 & 92.9 $\pm$ 0.04 \\
        ADA  & 28.7 $\pm$ 0.12 & 27.0 $\pm$ 0.08 & 76.5 $\pm$ 0.05 & 72.7 $\pm$ 0.05 & 95.2 $\pm$ 0.03 & 88.1 $\pm$ 0.04 \\ 
        ME-ADA & 29.4 $\pm$ 0.11 & 27.6 $\pm$ 0.11 & 77.1 $\pm$ 0.07 & \textbf{74.8 $\pm$ 0.03} & \textbf{95.6 $\pm$ 0.05} & \textbf{93.1 $\pm$ 0.04} \\ \hline
        RCAD & \textbf{30.4 $\pm$ 0.09} & \textbf{28.1 $\pm$ 0.10} & \textbf{77.3 $\pm$ 0.05} & 74.5$\pm$ 0.04 & \bf{95.7 $\pm$ 0.04} & 93.0$\pm$ 0.03
    \end{tabular}
    \caption{\textbf{Robustness to adversarial perturbations compared against in-distribution test accuracy:} Clean (evaluation on \iid test data) and Robust test accuracies of adversarial methods (adversarial training (FGSM~\cite{goodfellow2014explaining}), ADA and ME-ADA), ERM and RCAD when the adversary carries out attacks using the Fast Gradient Sign (FGSM) method when the strength of the attack in $l_1$ norm is $0.05$. We show the mean accuracy and $95\%$ confidence interval evaluated over $10$ independent runs.}
    \label{tab:compare-robust}
\end{table*}
\normalsize 

\subsection{Robustness to adversarial attacks} 

We first look at robustness to adversarial attacks, using FGSM~\citep{goodfellow2014explaining} as the attack method.
The conventional approach to fending off adversarial attacks is adversarial training, wherein the training objective exactly matches the testing objective. Thus, adversarial training represents the ``gold standard'' for performance on this evaluation.
We compare the adversarial robustness of RCAD, adversarial training, ADA, and ME-ADA in Table~\ref{tab:compare-robust}. Not only does our method achieve higher (clean) test accuracy than adversarial training on all datasets, but surprisingly it also achieves higher robust test accuracy on the harder CIFAR-100-2k benchmark where the clean test accuracy of RCAD is $+3\%$ greater than adversarial training, and robust test accuracy is $+0.5\%$ better than ME-ADA. Although, we find that adversarial baselines achieve better robust test accuracy when the training data is sufficient (\eg CIFAR-100 and CIFAR-10 in Table~\ref{tab:compare-robust}).

Both ADA and ME-ADA perform some form of adversarial training, so it is not surprising that they outperform RCAD on this task. We suspect that these methods outperform adversarial training because they are trained using the multi-step projected gradient descent (PGD)~\citep{madry2017towards}, rather than the one-step FGSM~\cite{goodfellow2014explaining} and also have a higher clean test accuracy. 
While our aim is \emph{not} to propose a state-of-the-art method for withstanding adversarial attacks, these preliminary results suggest that RCAD may be somewhat robust to adversarial attacks, but does so without inheriting the poor (clean) test accuracy of standard adversarial training.

\begin{figure}[!ht]
    \centering
    \includegraphics[width=0.7\linewidth]{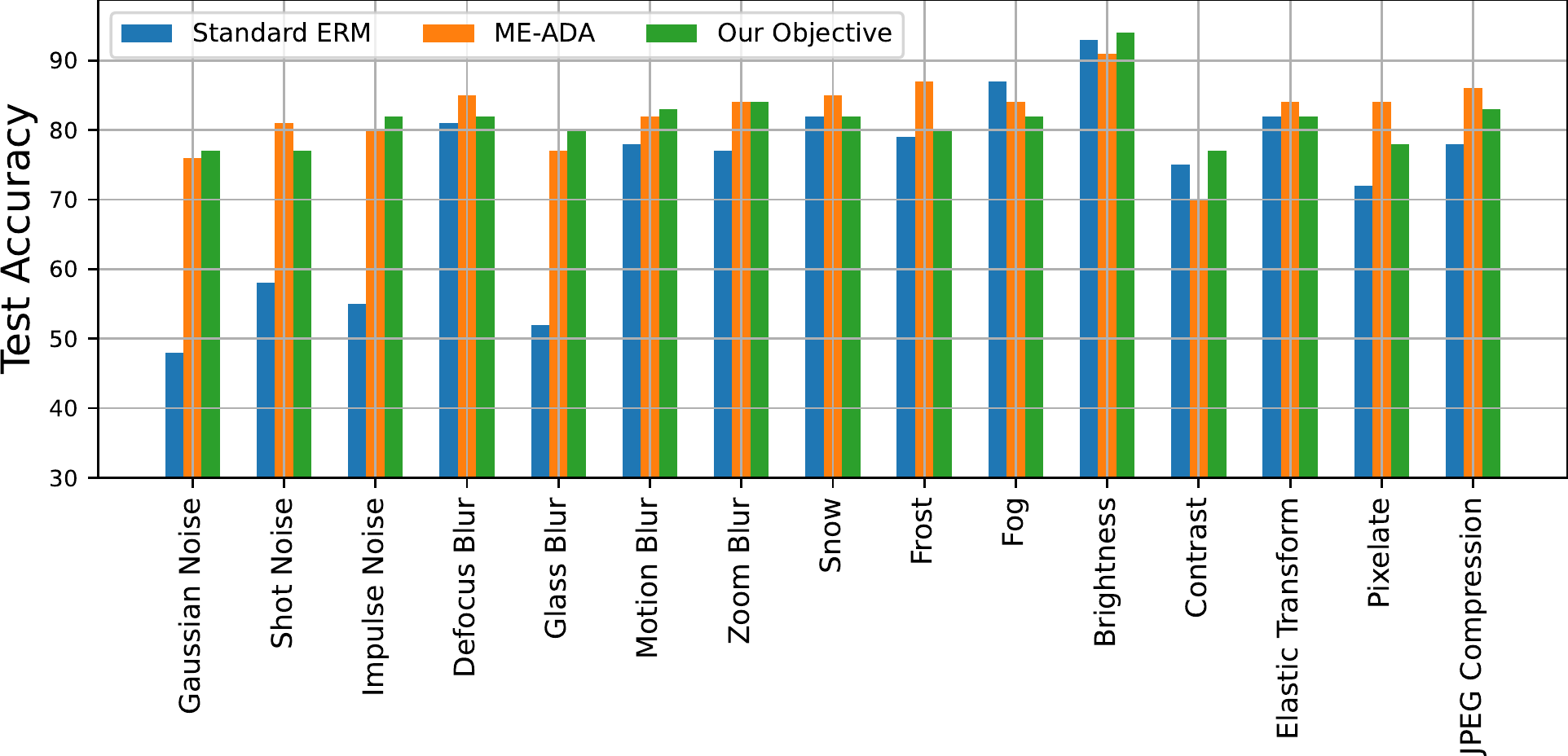}
    \caption{\textbf{Robustness to natural shifts in distribution:} Plots comparing the performance of standard ERM training and ME-ADA against our method (RCAD) trained on CIFAR-10 benchmark and tested on various distribution shifts in the corrupted CIFAR-10-C benchmark.}
    \label{fig:compare-cifar-c10}
\end{figure}

\subsection{Robustness to natural distribution shifts}

Our final set of experiments probe robustness to more systematic distribution shifts using the corrupted CIFAR-10-C dataset~\citep{hendrycks2019benchmarking}. These shifts go beyond the small perturbations introduced by adversarial examples, and are a more faithful reflection of the sorts of perturbations a machine learning model might face ``in the wild''. 

We compare RCAD to standard ERM and ME-ADA on this benchmark; when all methods are trained on the uncorrupted CIFAR-10 training dataset, but evaluated on different types of corruptions. We report results in Figure~\ref{fig:compare-cifar-c10}. Both RCAD and ME-ADA consistently outperform ERM. On certain corruptions (e.g., Gaussian noise, glass blur), RCAD and ME-ADA achieve test accuracies that are around $+25\%$ larger than the ERM baseline. We do not notice any systematic difference in the results of RCAD versus ME-ADA, but note that ME-ADA requires $2\times$ more compute than RCAD because its adversarial examples require multiple gradient steps to compute.

While the main aim of our experiments has been to show that RCAD achieves higher test accuracy, its good performance on robustness benchmarks suggests that it may be a simpler yet appealing choice for practitioners.

\section{Analyzing RCAD vs. ERM in a Non-linear Toy Classification Setup}
\label{app:sec:toy-non-linear}

In Section~\ref{sec:analysis} we analyse the solution returned by RCAD and compare it to the ERM solution in a simplified binary classification setting where the hypothesis class consists of linear predictors. We find that the high dimensional linear decision boundary learnt by ERM depends on the noisy dimensions that are spuriously correlated with the label on the training data $\hat \gD$. On the other hand, our method RCAD corrects this decision boundary by unlearning spurious features. The self-generated examples by RCAD exemplify the spurious correlations, and maximizing entropy on these examples helps the model to drive any learnt components along the noisy dimensions to lower values. In this section, we consider a setup where the true decision boundary is non-linear and the hypothesis class consists of two-layer neural networks with $128$ hidden units and ReLU activations. In order to induce spurious features in our training data, we stick to high-dimensional inputs.

Typically, in classification problems the true generalizable features span a lower dimensional space (compared to the ambient dimension) \citep{arjovsky2020invariant}.
Ideally, we would like to learn classifiers that depend only on these few generalizable features, and  remain independent of all the other noisy features.
However, training neural networks with the cross-entropy loss and stochastic gradient descent (SGD) often leads to overfitting on training data since the trained model ends up fitting on noisy  features with high probability~\cite{peters2016causal,heinze2017conditional}.
To simulate this phenomenon, we use a $d-$dimensional toy classification problem where the true features are given by the first two-dimensions only, while the rest of the $d-2$ dimensions are pure noise.

\begin{figure*}[t]
  \centering
  \begin{subfigure}[c]{0.23\textwidth}
    \setlength{\abovecaptionskip}{1pt} 
    \includegraphics[width=\linewidth]{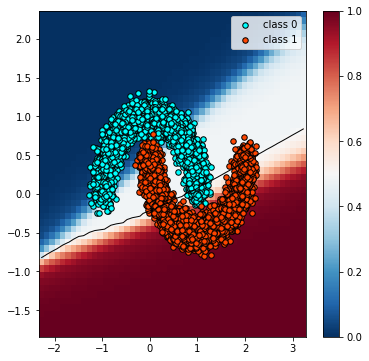}
    \caption{ERM \label{fig:toy-erm}}
   \end{subfigure}
  \begin{subfigure}[c]{0.23\textwidth}
    \setlength{\abovecaptionskip}{1pt} 
    \includegraphics[width=\linewidth]{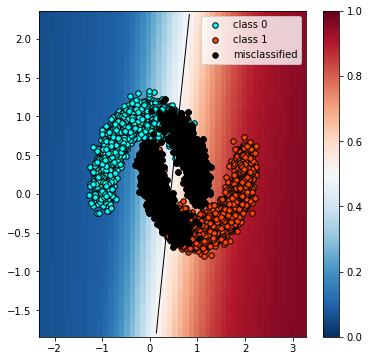}
    \caption{RCAD (100 epochs)}
   \end{subfigure}
  \begin{subfigure}[c]{0.23\textwidth}
    \setlength{\abovecaptionskip}{1pt} 
    \includegraphics[width=\linewidth]{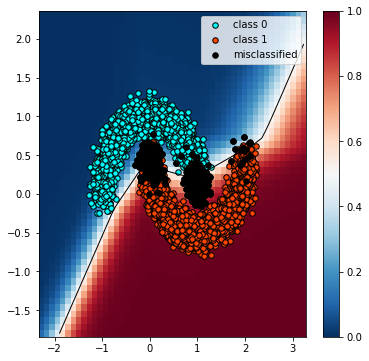}
    \caption{RCAD (200 epochs)}
   \end{subfigure}
  \begin{subfigure}[c]{0.23\textwidth}
    \setlength{\abovecaptionskip}{1pt} 
    \includegraphics[width=\linewidth]{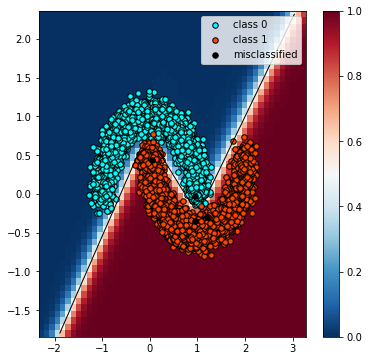}
    \caption{RCAD (300 epochs) \label{fig:toy-RCAD}}
   \end{subfigure}
  \caption{\textbf{RCAD objective learns decision boundary using only generalizable features:}
  We simulate a high-dimensional non-linear classification problem by projecting a simple 2-d dataset into a 625-dimensional space. \emph{(a)} Standard ERM training overfits to this dataset, achieving perfect training accuracy by picking up on spurious features. Plotting a 2D projection of the decision boundary along the true features, we see that it poorly separates the data along the generalizable (true) features.
    \emph{(b, c, d)} Visualizing our method (RCAD) at different snapshots throughout training, we see that it converges to the true decision boundary using only the generalizable features, thereby achieving a higher test accuracy.
}
  \label{fig:toy-data-results}
\end{figure*}

A dataset $\hat \gD$ of 20,000 training examples is generated by first sampling a two-dimensional vector $\tilde{\rvx}$ with equal probability from one of two classes supported over two different well separated moon shaped regions (see Figure~\ref{fig:toy-data-results}).  Here, $\tilde{\rvx}$ is a two dimensional vector with label $\ry \in \{0,1\}$. In our setting, the input dimension $d=625$. Now, in order to construct the final $d-$dimensional input $\rvx$ for the classifier using this two dimensional sample $\tilde{\rvx}$, we first append to each sample a vector of  ($d-2=623$) zeros, and then add add Gaussian noise $\rvepsilon \sim \calN(0, \sigma^2 \bI_d)$ where $\sigma=0.1$. Note that the noise is added to all dimensions, including the dimensions of the true feature but our results would not change if we choose to only add noise to the last $d-2$ dimensions, which would only make the problem easier.
\begin{align*}
    \rvx \triangleq (\tilde{\rvx}, \underbrace{0, \cdots, 0}_\text{d-2=623 zeros}) + \rvepsilon, \qquad \rvepsilon \sim \gN(\bf{0}, \sigma^2 \bf{I}_d)
\end{align*}

\begin{table}[!t]
    \centering
    \footnotesize
\begin{tabular}{rccc|c}
            $(d, \sigma)$ & ERM & Adv. Training (FGSM) & ME-ADA & RCAD  \\ \hline
            $(625,0.10)$ 
            & $80.4\%$ & $78.8\%$ & $90.9\%$ & $\mathbf{94.9\%}$ \\ 
            $(625,0.75)$
            & $77.8\%$ & $77.9\%$ & $84.4\%$ & $\mathbf{89.6\%}$ \\ 
            $(1000,0.10)$
            & $74.3\%$ & $72.0\%$ & $83.5\%$ & $\mathbf{87.1\%}$ 
        \end{tabular}
        \vspace{0.1in}
    \caption{\textbf{In the toy non-linear classification problem, RCAD is found to be more robust to spurious correlations in the training data.} We show the test performance of RCAD vs. baseline methods ERM, adversarial training (FGSM) and ME-ADA as we vary two parameters: $d$ which is the dimension of the input, and $\sigma$ which is the standard deviation of the noise added to every dimension. \vspace{-0.2in}}
    \label{tab:toy-nonlinear}
\end{table}
        
We will measure the in-distribution performance of a classifier, trained on $\hat \gD$ and evaluated on fresh examples from the  distribution defined above.
In our toy setup, the first two dimensions of the data perfectly explain the class label; we call these dimensions the generalizable (true) features. We are interested in learning non-linear classifiers (two-layer neural nets) that successfully identify these generalizable features, while ignoring the remaining dimensions.

Models trained with the standard cross-entropy loss on a fixed dataset $\hat \gD$ are liable to overfit~\citep{zhang2021understanding}. A model can best reduce the (empirical) cross-entropy loss by learning features that span all dimensions, including the spurious feature dimensions. Precisely, the model can continue to reduce the cross-entropy loss on some example $\rvx^{(i)}$ by further aligning some of the weights of its hidden units along the direction of the corresponding noise $\rvepsilon^{(i)}$, that is part of input $\rvx^{(i)}$.

We start by training a two-layer net on $\hat \gD$. We train it for 300 epochs of SGD using the standard empirical cross entropy loss. This model achieves perfect training accuracy ($100\%$), but performs poorly on the validation set ($80.4\%$). To visualize the learned model, we project the decision boundary on the first two coordinates, the only ones that are truly correlated with the label on population data.
The decision boundary for this model, shown in Figure~\ref{fig:toy-erm}, is quite different from the true decision boundary. Rather than identifying the true generalizable features, this model has overfit to the noisy dimensions, which are perpendicular to the span of the true features.
Training this model for more epochs leads to additional overfitting, further decreasing the test accuracy.

Next we train with our objective RCAD using the same training dataset. In addition to minimizing the standard empirical cross entropy loss, our method also maximizes the predictive entropy of the model on self-generated perturbations that lie along the adversarial directions. Intuitively, we expect that these new examples will be along the directions of the spurious features, exacerbating them. This is because adversarial directions have been shown to comprise of spurious features~\cite{ilyas2019adversarial}. Thus, in training the model to be less confident on these examples, we signal the model to unlearn these spurious features.
Applying RCAD to this dataset, we achieve a much larger test accuracy of $94.9\%$. When we visualize the decision boundary (along span of the true features) in Figure~\ref{fig:toy-RCAD}, we observe that it correctly separates the data.
While SGD is implicitly biased towards learning simple (e.g., linear) decision boundaries~\citep{kalimeris2019sgd}, our results show that RCAD partially counters this bias, 
forcing the model to learn a non-linear decision boundary along the true features and ignoring the noisy dimensions. Additionally we also train adversarial baselines in this setup and compare the test accuracies with RCAD in Table~\ref{tab:toy-nonlinear}. We find that RCAD is less vulnerable to spurious correlations compared to other methods even when we increase the dimensionality of the input or the noise level in the data generating process, both of which would make models more prone to overfitting. This is confirmed by the significant drop in the test performance of the ERM baseline.

\end{document}